\pdfoutput=1

\documentclass[11pt]{article}

\usepackage{acl}

\usepackage{times}
\usepackage{latexsym}
\usepackage{subfigure}
\usepackage{multicol}
\usepackage{hyperref}
\usepackage{placeins}

\usepackage{graphicx}
\usepackage{bera}
\usepackage{pgfplots}
\pgfplotsset{compat=newest,compat/show suggested version=false}

\usepackage{skak}

\usepackage{booktabs,dcolumn}
\usepackage{xcolor}
\definecolor{darkgreen}{RGB}{0, 153, 51}
\definecolor{darkred}{RGB}{204, 51, 0}

\usepackage{tikz}
\usetikzlibrary{patterns}
\usepackage{comment}
\usepackage{comment}

\usepackage{ntheorem}
\theoremseparator{:}
\newtheorem{hyp}{Hypothesis}

\usepackage[T1]{fontenc}
\usepackage{enumitem}

\usepackage[utf8]{inputenc}

\usepackage{microtype}

\title{Comparing informativeness of an NLG chatbot vs graphical app in diet-information domain}

\author{Simone Balloccu \\
  University of Aberdeen, UK \\
  \texttt{simone.balloccu@abdn.ac.uk} \\\And
  Ehud Reiter \\
  University of Aberdeen, UK \\
  \texttt{e.reiter@abdn.ac.uk} \\}

\begin{document}
\maketitle

\begin{abstract}
Visual representation of data like charts and tables can be challenging to understand for readers. Previous work showed that combining visualisations with text can improve the communication of insights in static contexts, but little is known about interactive ones. In this work, we present an NLG chatbot that processes natural language queries and provides insights through a combination of charts and text. We apply it to nutrition, a domain communication quality is critical. Through crowd-sourced evaluation, we compare the informativeness of our chatbot against traditional, static diet apps. We find that the conversational context significantly improved users' understanding of dietary data in various tasks and that users considered the chatbot more useful and quick to use than traditional apps.
\end{abstract}

\begin{figure*}[t]
\centering
    \includegraphics[scale=0.37]{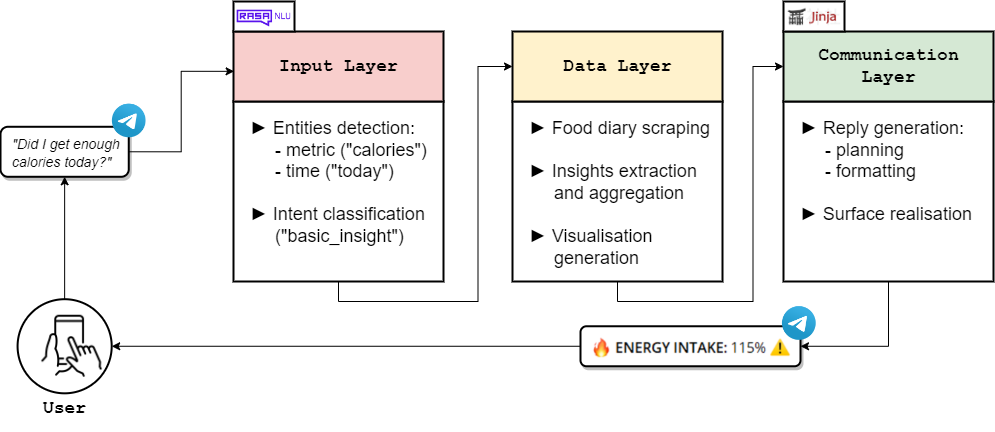}
    \caption{Chatbot architecture and interaction flow.}
    \label{fig:arch}
\end{figure*}

\section{Introduction}
Visual representations of data are commonly used to communicate insights to the reader. However, understanding the meaning of charts or other visualisations can be challenged by visual deficit, information context, or just the required cognitive effort. Previous research investigated on generating  textual explanations of data and comparing them with visualisations \cite{gatt2009data,molina2011generating,gkatzia2017data}.
Approaches like these are particularly useful in healthcare, where lots of data get produced and communication plays a critical role \cite{zolnierek2009physician,brock2013republished}. 
Most of these works showed that combining text and visuals improves users' understanding of data but they explored static contexts only, where information is presented in a fixed way and there is no active interaction with the reader. Little is known about the effects of text and charts combination in dynamic scenarios, such as conversational ones. Since chatbots are emerging as tools for healthcare \cite{zhang2020artificial}, it is important to assess if they can provide better communication than static tools (e.g. e-health apps).

In this work, we develop and evaluate an NLG-chatbot that generates insights explanations by combining graphics and text. Using our chatbot, users do not need to explore or interpret data themselves, as they can directly ask what they're looking for and get it, along with an explanation. We apply it to diet coaching, a domain where communication quality is critical \cite{van2008cognitive,savolainen2010dietary,michie2011behaviour} and often overlooked by existing tools \cite{10.1145/3450614.3463602,balloccu-reiter-2022-beyond}. To assess the effectiveness of this approach, we run a human evaluation in which we compare our chatbot with traditional diet apps. Participants were assigned to either our chatbot or an app and used it to take a 10-point quiz concerning the extraction of insights from a simulated food diary. In the end, participants expressed feedback on the assigned tool. Results show that using our chatbot led to significantly higher scores compared to using traditional apps, both in general and with regard to particular sub-topics. Feedback analysis also reveal that participants perceived our chatbot as more useful for finding diet problems and quicker to use than traditional diet apps. 

\section{Related work}
In this section, we recap past research on charts and text combination for insight explanation. We first look at more general work, then move to healthcare and diet-coaching.

\subsection{Text vs Graphics in NLG}
Previous work investigated how NLG can enhance the understanding of data by combining textual content and images. Work on weather data \cite{gkatzia2017data}, showed mixed text and pictures improving decision-making over images alone. Dashboards \cite{ramos2017evaluation} benefit from textual explanation of charts as well, as it helps assess learning in students. Combining charts with explanation of sensor data \cite{molina2011generating} helps insight understanding for general users. Driving reports \cite{braun-etal-2015-creating} are more helpful if presented as a mix of pictures and text. Healthcare data can also be explained through NLG \cite{Pauws2019}. Experiments in NICU \cite{law2005comparison,van2010graph} suggest that combining charts and text could be the preferred approach by clinicians.

\begin{figure*}[t]
\centering
    \includegraphics[scale=0.37]{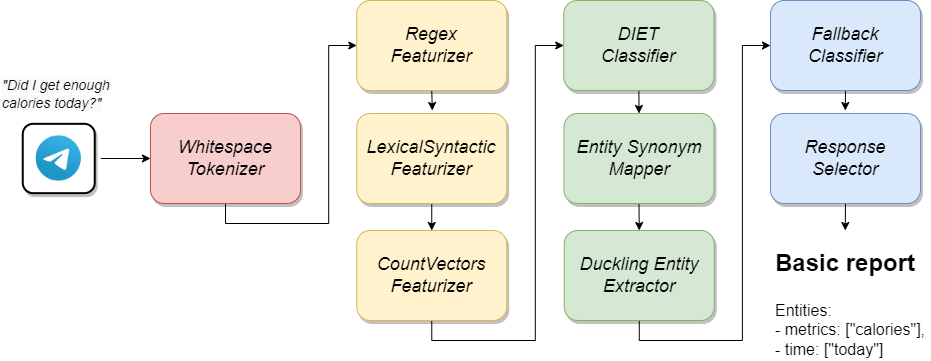}
    \caption{Overview of the NLU pipeline.}
    \label{fig:nlu}
\end{figure*}

\subsection{Text vs Graphics in diet-coaching}
Information quality and communication play a big role in diet \cite{van2008cognitive,savolainen2010dietary,michie2011behaviour}. This applies to apps as well: comprehensibility showed to be a predictor of prolonged app use \cite{lee2017motivates}. Sub-optimal communication can confuse and demotivate users, leading to early abandonment \cite{murnane2015mobile, Mukhtar2016}.
Despite this, diet apps (like MyFitnessPal \footnote{www.myfitnesspal.com} or FatSecret\footnote{https://www.fatsecret.com/}) typically come as calorie counters, where users log their meals to obtain insights. These tools adopt very limited textual communication and make extensive use of visualisations that must be interpreted by users themselves \cite{balloccu-reiter-2022-beyond}. Considering the relationship between numeracy and nutrition literacy \cite{mulders2018label}, this poses a barrier between users and the delivered information. Our previous work \cite{10.1145/3450614.3463602} showed similar issues for conversational agents: chatbots adopt fixed educational material \cite{10.1145/3267305.3274191,stephens2019feasibility,davis2020process}, such as PDFs containing guidelines, and expose a lack of reasoning over user queries \cite{maher2020physical}. Similarly to apps, chatbots show plain reports, with little to no feedback on goals, progress or mistakes \cite{10.1145/3267305.3274191,10.1145/3421937.3421960}.

\begin{figure}[ht]
    \centering
    \includegraphics[scale=0.5]{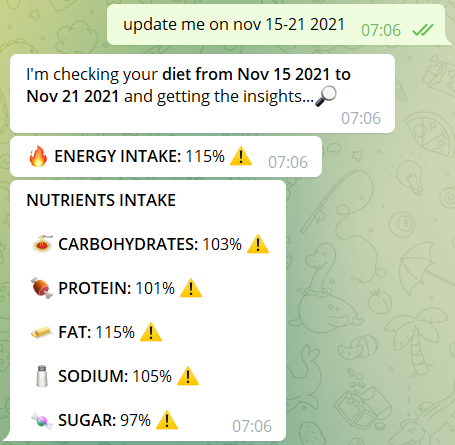}\\
    \vspace{1mm}
    \includegraphics[scale=0.5]{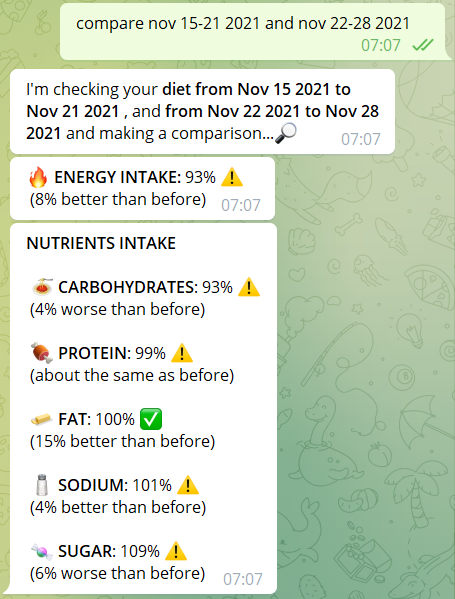}
    \caption{Basic report and comparison as presented by the chatbot.}
    \label{fig:basic_and_compare}
\end{figure}

\section{NLG chatbot to improve communication quality}
Our chatbot consists of an Input Layer for users' input understanding; a Data Layer that extracts insights and generates visualisations; a Communication Layer that performs planning and surface realisation (\autoref{fig:arch}). We use RASA Open Source 2.0\footnote{https://rasa.com/docs/rasa/} as the main infrastructure for the entire system, and exploit its NLU component (\autoref{fig:nlu}) for the Input layer; the Data Layer adopts a custom data analysis logic; the Communication Layer adopts rule-based NLG and variable templates (through Jinja 3.0 \footnote{https://jinja.palletsprojects.com/en/3.0.x/}).

We adopt a hybrid architecture: we use machine-learning for NLU but restrict text generation to rules. This is mainly for two reasons: 1) diet domain imposes strict accuracy requirements that cannot be met by current E2E NLG \cite{thomson-reiter-2020-gold,van-miltenburg-etal-2021-underreporting} and 2) to the best of our knowledge, there is no publicly available diet-coaching corpus which can be used to train or fine-tune generative models. On the other hand, machine-learning offers good generalisation for NLU with the only risk being unexpected inputs or failure in intent classification. 

We model two main interactions in the chatbot: basic reports and comparisons (\autoref{fig:basic_and_compare}). Basic reports show insights about a single time frame, either as brief information on energy and nutrients balance or combinations of charts and text. Comparisons extend basic reports to multiple time frames by informing users about progress (e.g. improved intake; changes in food choices etc..). For each request, users can specify metrics (calories and five nutrients: carbohydrates, protein, fat, sugar and sodium) and time (detected via Duckling Entity extractor\footnote{https://duckling.wit.ai/}). This approach offers more flexibility than traditional apps, which typically aggregate all the metrics in a single section (e.g. a table) and present pre-defined comparisons (e.g. every month).

\begin{figure}[ht]
    \centering
    \includegraphics[scale=0.5]{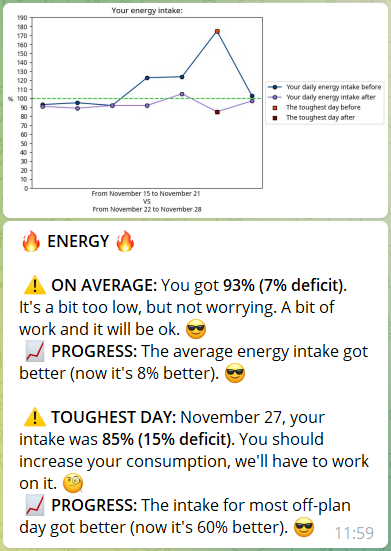}\\
    \vspace{1mm}
    \includegraphics[scale=0.5]{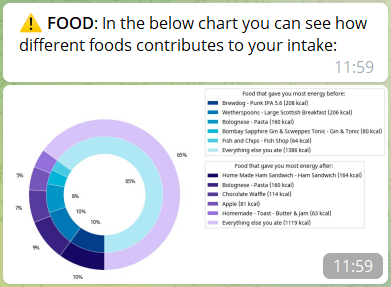}
    \caption{Example of advanced insights (intake and food analysis) for comparisons.}
    \label{fig:adv_comp}
\end{figure}

\subsection{Explanation through text and charts}
\begin{figure}[ht]
    \centering
    \includegraphics[scale=0.5]{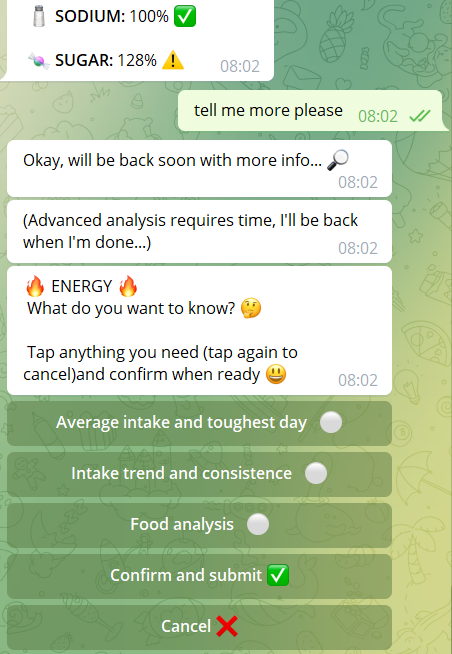}\\
    \vspace{1mm}
    \includegraphics[scale=0.44]{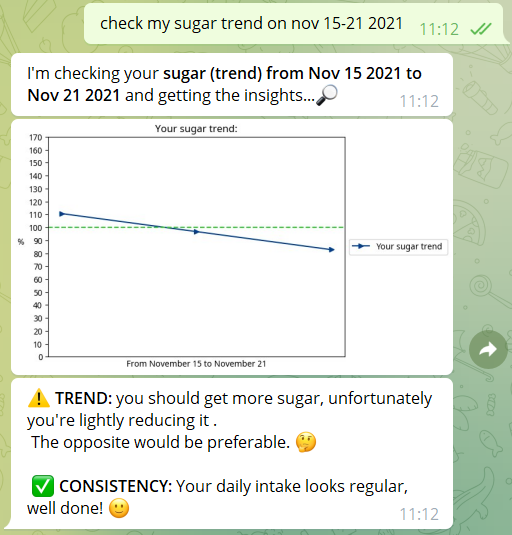}
    \caption{Obtaining advanced insights: Guided navigation with buttons (top) VS Natural language query about trend and consistency (bottom).}
    \label{fig:guided_query}
\end{figure}

Users can access two typologies of insights: basic and advanced. Basic insights show energy and nutrient intake (see \autoref{fig:basic_and_compare}) as brief textual messages. This is thought for users that need a quick glance at their data. Advanced insights deliver more information and are presented as a combination of text and charts. Users can obtain the following advanced insights (\autoref{fig:adv_comp}):
\begin{enumerate}
    \item \textbf{Intake analysis:} reasons and explains intakes with regards to user goals.
    \item \textbf{Trend and consistency:} detects if trends match recommended changes in diet (e.g. getting fewer calories to fix an excess) and checks intake consistency (maintaining a stable intake across days).
    \item \textbf{Food analysis:} reasons and explains intakes at the food level, by showing which food has the biggest impact.
\end{enumerate}
Advanced insights naturally extend to comparisons as well (\autoref{fig:adv_comp}). To let both novice users (that need supervision) and advanced ones access advanced insights, they can be obtained in two ways (\autoref{fig:guided_query}): 
\begin{enumerate}
    \item \textbf{Guided navigation:} through generic queries (e.g. "tell me more about this" or "anything else?"). Following this trigger, the chatbot presents a button interface for each available advanced insight. Buttons can be checked and unchecked to obtain only those insights that are of interest.
    \item \textbf{Natural language query:} by directly asking for specific insights and metrics. This can be done by specifying a particular insight (e.g. "food" or "intake") on a specific period.
\end{enumerate}
For both interactions, users can specify one or more metrics.

\subsection{Other features}
We implement a number of supplementary best practices \cite{ferman2018towards} to further improve usability and clarity. The chatbot actively provides feedback for each input (while informing users on the pending task); adopts emojis to make insights more understandable; splits the content into multiple messages and introduces a dynamic delay between them to avoid flooding.

\begin{figure*}[ht]
\centering
    \includegraphics[scale=0.31]{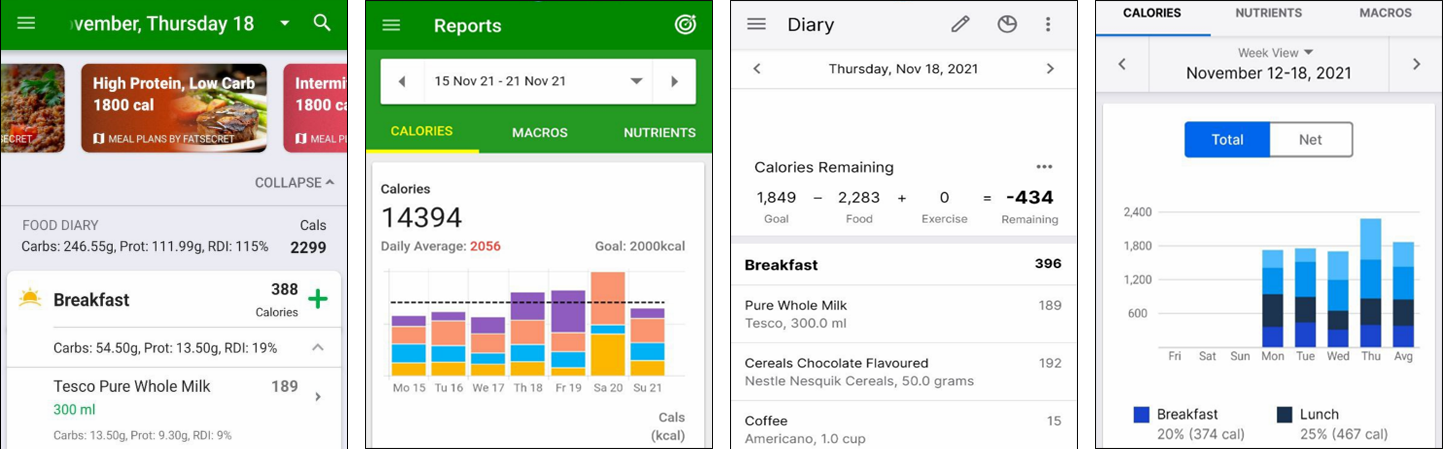} 
    \caption{Food Diary and nutrition reports as shown to the user in FatSecret (left) and MyFitnessPal UI (right).}
    \label{fig:mfpfs}
\end{figure*}

\section{Experiment setup}
We deploy our chatbot on Telegram Bot API\footnote{https://core.telegram.org/bots/api} and compare its informativeness with traditional diet apps. We gather our test population (\textbf{workers}) through crowd-sourcing on Amazon Mechanical Turk\footnote{https://www.mturk.com/worker/help}. Details of recruitment, pay and sanity checks are available in the Appendix\ref{sec:ethics}. We choose to compare our chatbot with MyfitnessPal\footnote{https://www.myfitnesspal.com/} (\textbf{MFP}) and FatSecret\footnote{https://www.fatsecret.com/} (\textbf{FS}). An example of the two apps UI can be seen in \autoref{fig:mfpfs}. We choose these two apps based on their popularity and downloads number on the Apple and Android app stores. We do not compare against any dieting chatbot as none of those present in the literature is publicly available.

\subsection{Measuring informativeness}
Aiming at communication improvement, we need to find a measure to capture whether one specific tool performs better than others. From communication theory, \cite{webster2019communicative} we adopt the concept of "informativeness", defined as "how successfully a person is able to convey an intended message". We extend this definition to diet systems as "how successfully a tool is able to convey an intended message". To capture informativeness we create a ten questions quiz regarding diet analysis (a sample is provided in Appendix \ref{sec:appendix}). The quiz consists of 4 macro-tasks:
\begin{enumerate}
    \item \textbf{Day analysis:} understanding if calories and carbohydrates are balanced on a single day (2pts).
    \item \textbf{Food analysis:} understanding what food provided the most calories and fat on a single day, along with quantities (4pts).
    \item \textbf{Week analysis:} understanding if calories and carbohydrates are balanced across a week (2pts).
    \item \textbf{Weeks comparison:} understanding if, by comparing two weeks, calories and carbohydrates improved or worsened (2pts).
\end{enumerate}
Each question is worth 1 point, for a total of 10 points. We choose to develop a custom quiz because no available questionnaire can be used to evaluate the informativeness of a diet-coaching tool. In creating it, we analyse existing apps and all the information that they deliver; we incorporate experts' recommendations from previous surveys \cite{nu12082214}; we consider the theoretical constructs of self-regulation \cite{zahry2016content}, with a particular focus on the measure of informativeness. We avoid evaluating the "trend and consistency" feature for fairness, as apps don't offer a way for the user to infer such information without long and tedious calculations. 
\par Workers were randomly assigned to either our chatbot, MFP or FS, each of which was pre-filled with a simulated food diary (none of the data belonged to the users) consisting of 2 weeks of logged meals. We obtained n=27 workers assigned to our chatbot; n=31 workers to MFP; n=29 workers to FS. Besides the tool itself, workers were provided with a PDF guide on how to use it and a glossary explaining the meaning of the terms used in the quiz. Each worker took the quiz and was asked to answer the questions to the best of their knowledge by using the tool. Through the quiz we test the following hypothesis:
\begin{hyp}[H\ref{hyp:1}] \label{hyp:1}
Chatbot workers scored higher on the informativeness quiz than MFP or FS workers.
\end{hyp}

\begin{table*}[t]
    \centering
    \vspace{-9mm}
    \resizebox{310pt}{!}{
    \hspace{-4mm}
    \begin{tabular}{|c|c|c|c|c|c|c|}
    \cline{2-7} 
    \multicolumn{1}{c|}{} & \multicolumn{3}{c|}{\bf Average score} & \multicolumn{3}{c|}{\textbf{Score differences}} \\ \hline
    \textbf{Topic} &  CB & FS & MFP &  CB-FS & CB-MFP & MFP-FS \\ \hline
    Overall (10pt) & \textbf{6.65 }  & 4.13 & 5.22 & \color{darkgreen} +2.52** & \color{darkgreen} +1.43 & \color{darkgreen} +1.09\\ \hline
    Day analysis (2pt) & 1.15 & 0.76 & \textbf{1.32}   & \color{darkgreen} +0.40 & \color{darkred} -0.16 & \color{darkgreen} +0.56  \\  \hline
    Food analysis (4pt) & \textbf{2.85} & 2.14 & 0.91 & \color{darkgreen} +0.71 & \color{darkgreen} +1.94*** & \color{darkred} -1.23* \\ \hline 
    Week analysis (2pt) & \textbf{1.35}  & 0.66 & 1.05 & \color{darkgreen} +0.70**  & \color{darkgreen} +0.30 & \color{darkgreen} +0.39 \\ \hline 
    Weeks comparison (2pt) & \textbf{1.31} & 0.59 & 1.14 & \color{darkgreen} +0.72** & \color{darkgreen} +0.17 & \color{darkgreen} +0.55** \\ \hline
    \end{tabular}}
      \caption{Results from informativeness quiz. On the left side: average scores, overall and for specific tasks. The highest score for each category is in bold. On the right side: score differences between tools. Green is for higher scores, and red is for lower scores. CB = Chatbot; MFP = MyFitnessPal; FS = FatSecret. Significance: * for p<0.05; ** for p<0.01; *** for p<0.001.}
      \label{tab:quiz}
\end{table*}

\subsection{Measuring nutrition literacy}
Previous research highlighted the importance of nutrition literacy in dieting \cite{michie2011behaviour}, so we analyse its impact on our experiment. We also analyse if our chatbot communication can reduce the score gap between different literacy levels. Before taking the quiz, each worker completed Pfizer's Newest-Vital-Sign ({\bf NVS}) \cite{weiss2005quick,powers2010can}, consisting of 6 questions (each one worth 1 point) regarding an ice-cream label. NVS scores are grouped in ranges: 0-1 refers to "high likelihood of limited literacy", 2-3 refers to "possibility of limited literacy"; 4-6 refers to "adequate literacy". We compare NVS scores with quiz scores to test the following hypothesis:
\begin{hyp}[H\ref{hyp:2}] \label{hyp:2}
There was a positive correlation between NVS score and quiz score in our experiment, but not for chatbot workers.
\end{hyp}

\subsection{Measuring perception of the tool and past experience}
Finally, we inspect workers' opinions on the tool they used. We ask each worker to rate the tool under different characteristics (see \autoref{fig:likert}) through Likert-5 scale.
Through this approach we test the following hypothesis:
\begin{hyp}[H\ref{hyp:3}] \label{hyp:3}
Our chatbot received higher ratings across the proposed questions.
\end{hyp}
Finally, we ask workers to specify whether they had past experience with dieting tools (including the one they were assigned) and to specify how often they used them (often; occasionally; rarely; never).

\section{Results analysis}
For variance analysis, we adopt One-Way ANOVA and Tukey's post-hoc test (replaced respectively by Kruskal-Wallis test and Dunn's post-hoc test if ANOVA's normality requirement is not met). To test variable dependence we adopt Chi-squared test and Bonferroni's post-hoc test. For the correlation test, we adopt Pearson correlation (substituted by Spearman correlation if Pearson's normality requirement is not met).

\subsection{Preliminary checks}
Before analysing results, we verify nutrition literacy uniformity across our population, to ensure that none of the groups contained mostly workers with high/low nutrition literacy. We discover that nutrition literacy distribution is unbalanced among apps, with the majority of workers with low nutrition literacy assigned to MFP sample, none to FS only one to our chatbot (see \autoref{tab:nvs}). We re-balance the samples by removing all such workers. This limits our inspections of nutrition literacy but keeps the comparison fair. From now on, all results will refer to the re-balanced sample unless otherwise specified. We also check for meaningful differences in workers' past experience with diet tools, but find none neither in general ($p=0.47$) and by considering only those workers who had past experience and ($p=0.27$).

\begin{figure*}[ht]
    \centering
    \vspace{-5mm}
    \includegraphics[scale=1.5,trim={0 0cm 11.2cm 0},clip]{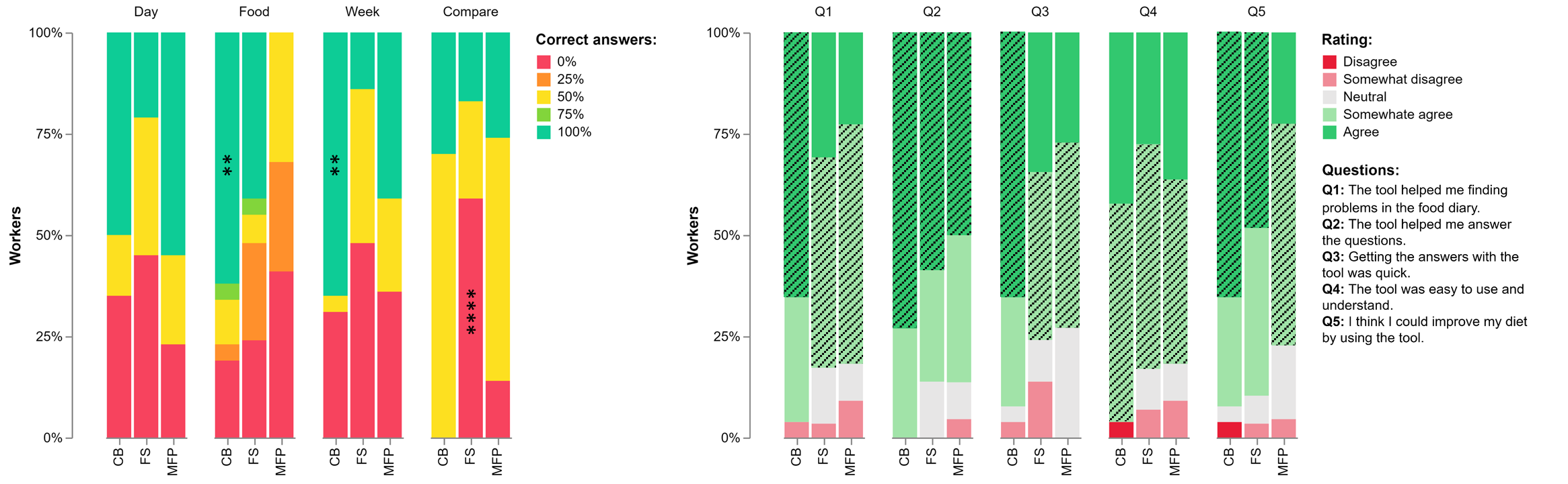}\\
    \caption{Percentage of correct answers by task, for each tool. CB = Chatbot; FS = FatSecret; MFP = MyFitnessPal. For day, week analysis and comparisons (2-points) we check no right answer (0\%), 1 right answer out of 2 (50\%) and all right answers (100\%). For food analysis (4 points), we check quarters as well. Significance: * for p<0.05; ** for p<0.01; *** for p<0.001; **** for p < 0.0001.}
    \label{fig:completion_rate}
\end{figure*}

\begin{table}
    \centering
    \resizebox{135pt}{!}{
    \vspace{1mm}
    \hspace{-4mm}
    \begin{tabular}{|c|c|c|c|}
    \cline{2-4} 
    \multicolumn{1}{c|}{} & \multicolumn{3}{c|}{\bf Workers per class} \\ \hline
    \textbf{NVS class} &  CB & FS & MFP \\ \hline
    LOW (0-1pt)  & 1  & 0 & 9\\ \hline
    MID (2-3pt) & 5 & 3 & 5  \\  \hline
    HIGH (4-6pt) & 21 & 26 & 17 \\ \hline 

    \end{tabular}}
      \caption{Distribution of nutrition literacy for our population. CB = Chatbot; MFP = MyFitnessPal; FS = FatSecret.}
      \label{tab:nvs}
\end{table}

\subsection{Quiz scores}
We first check total and per-task quiz scores (see \autoref{tab:quiz}). We find that, overall, the highest average score was reached by chatbot workers. The difference was statistically significant when compared to FS workers. Regardless of the group, average scores were low, not going much higher than 6/10. We consider this as a further confirmation of how hard understanding dietary insights is for the average user, especially in our context where data was simulated. By inspecting individual quiz tasks, we see that chatbot workers scored significantly higher in week analysis and comparison than FS workers, and in food analysis than MFP workers. We also find that MFP workers scored significantly higher than FS workers when comparing weeks, while the opposite happened for food analysis. Overall, chatbot workers always scored the highest score in every case, except for the day analysis, where MFP workers' scores were slightly higher.
\par Next, we look at the percentage of correct answers to check if any of the tools were associated with reaching specific scores (e.g. maximum points or 0 points). First, we find that our chatbot was positively associated ($p=0.0001$) with an overall score of 9/10 points. This tells us that the chatbot made it easier to reach higher scores in general. We then proceed to analyse individual quiz tasks (\autoref{fig:completion_rate}). Our chatbot was positively associated with the maximum score in food analysis and week analysis. For chatbot workers, it was easier to understand food details and insights based on aggregation in general. It was also negatively associated ($p=0.001$) with 0 points in weeks comparison. In fact, every chatbot worker managed to answer at least one of the two questions about comparison right. Interestingly, we find the opposite for FS, which was positively associated with scoring 0 points in weeks comparison. This tells us that FS workers struggled considerably in this task. Lastly, using MFP was negatively associated with the maximum score in food analysis: understanding food details was one of the hardest tasks with MFP.

\subsection{Nutrition Literacy effect on scores}
We check if nutrition literacy influenced quiz scores. Here we discover a discrepancy between the balanced and unbalanced samples. MFP workers show a significant difference ($p=0.03$) in scores between high and low nutrition literacy. By re-balancing the sample, we lose this significance. We also discover a moderate correlation ($\rho = 0.48, p=0.02$) between nutrition literacy and quiz score for MFP workers, even after balancing the samples.

\begin{figure*}[!ht]
    \centering
    \vspace{-5mm}
    \includegraphics[scale=1.5,trim={8.4cm 0cm 0 0},clip]{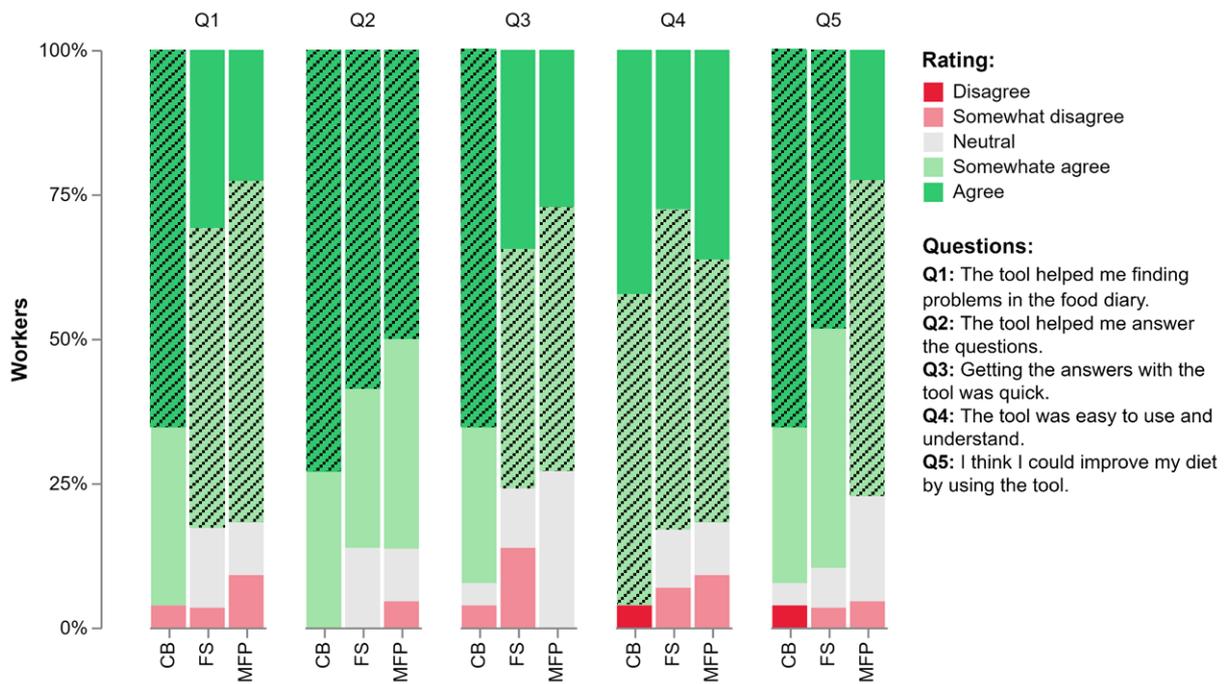}\\
    \caption{Feedback from users, based on used tool. CB = Chatbot; FS = FatSecret; MFP = MyFitnessPal. Lined bars indicate the mode for each question.}
    \label{fig:likert}
\end{figure*}

\subsection{Users' perception of the tool}
Finally, we check workers' feedback (see \autoref{fig:likert}). We notice a generally positive evaluation for every tool, with the chatbot getting a higher amount of "Agree" ratings across every question. By single-item analysis, our chatbot was positively associated with "Agree" in Q1 ($p=0.01$), where it also shows a better mode value than the other tools. Chatbot workers felt it more useful for finding problems in the food diary. We also find a better mode than both apps in Q3, meaning that workers found it to be quicker to use. This result in particular is unexpected considering that there was no significant difference in the quiz execution time ($p=0.22$). It could be that using natural language in our chatbot was felt as faster than navigating through different app sections. No app showed better mode than our chatbot in any question. Finally, it is interesting to notice that FS scored higher than MFP in Q5 despite being the tool with the lowest scores across every task except food analysis.

\section{Discussion}
From quiz results, chatbot workers scored the highest in informativeness across every scenario except for a slight advantage of MFP in day analysis. In multiple contexts, the difference between MFP and FS was statistically significant. We also found that using the chatbot was associated with a higher completion rate in different tasks and very high overall scores like 9/10. With these results we confirm H\ref{hyp:1}. We could not inspect nutrition literacy properly, as the different samples were too unbalanced and introducing low-literate workers would have made the comparison between MFP and our chatbot unfair. We saw a relationship between lower nutrition literacy and quiz scores, but isolated to MFP workers, and could not verify it across the whole population. With these results, we neither confirm nor reject H\ref{hyp:2} because of the lack of data. Looking at the feedback, we found out that our chatbot received a higher amount of "Agree" ratings across every question. It was also the only tool that showed association with maximum usefulness in finding diet problems. By analysing the mode of each question, we discovered that our chatbot was evaluated as quicker to use than the other apps. We also see that, unlike MFP and FS, it never showed a lower mode than any other tool. With these results we confirm H\ref{hyp:3}.

\section{Conclusion and future developments}
In this work, we evaluated the combination of charts and textual explanations for diet coaching, in the conversational scenario. We implemented an NLG-chatbot that understands natural language input and returns dietary insights as a combination of textual explanations and visualisations. We compared the chatbot with traditional static diet apps by inspecting informativeness and user feedback. Results show that the combination of visuals and text efficiently delivers information in diet-coaching, and makes it more understandable. Improved informativeness could play a critical role in diet outcomes. Feedback was generally more positive for the chatbot, meaning that it can be a valid tool for diet-coaching, potentially substituting static apps.

For future work, we plan to investigate if our approach can lead to actual learning from the user, for example through spaced repetition \cite{ausubel1965effect,doi:10.1073/pnas.1815156116} that can positively affect users' forgetting curve \cite{ebbinghaus2013memory}. We also commit to addressing the limits of our setup, to properly inspect the relationship between nutrition literacy and informativeness. We also plan to inspect more personalised approaches to information tailoring, namely by considering users' stress and emotional state that showed to be promising research directions \cite{balloccu-etal-2020-introducing,balloccu-reiter-2022-beyond}. Lastly, we consider this result as a sign of the maturity of our approach and we plan to run a trial to measure its effect on diet-coaching (e.g. weight control).

\bibliography{anthology,custom}
\newpage
\onecolumn
\newpage
\twocolumn
\appendix
\section{Ethics}
\label{sec:ethics}
This section sums up the procedure we adopted to ensure the ethical compliance of our experiment.

\subsection{Preliminary review}
Before starting the experiment, the procedure and materials were carefully reviewed by the University of Aberdeen's Ethics Board. Our experiment proposal was accepted without major revisions.

\subsection{Platforms}
For the quiz, we adopted Microsoft Forms\footnote{https://forms.office.com/} because of its compliance with GDPR policy. For hiring, we used Amazon Mechanical Turk. No recruitment qualification was specified, besides custom ones to prevent the same worker from submitting multiple HITs. Participants were shown a consent form containing all the information regarding the experiment procedure. They were also informed about the requirements that had to be satisfied to obtain the remuneration. All workers had to confirm their acceptance of these conditions (through checkboxes) in order to proceed with the experiment. Workers were given an email contact in case of problems during the experiment.

\subsection{Pay and workload}
Before launching the experiment, we verified the average completion time with 10 test users. The average result for completing the whole experiment (reading information; downloading and setting up material; taking NVS; taking the quiz; expressing the feedback) was 20 minutes. We gave each worker 45 minutes and paid 15 USD for the HIT. Workers were informed that if they ran out of time on Mturk they could just finish the quiz (on Microsoft Forms web platform) and contact us through the provided email address to still get paid.

\subsection{HITs sanity checks}
We received a total amount of 250 applications for our task. Most of them were fraudulent, with random answers or unrealistic completion times. In order to recognise legit HITs, we set up multiple sanity-checks, both in general and depending on the tool each worker was assigned.

\subsubsection{Global sanity checks}
To check the attention of workers during Pfizer's NVS, a fake price was added to the ice-cream label. Consequently, we added a (non-scored) question to the form, asking "what's the price of the ice-cream?". Moreover, each worker received a completion code that they had to submit on the Mechanical Turk platform after completing all the tasks.

\subsubsection{Sanity check for chatbot worker}
The chatbot was programmed to accept some custom queries that led to specific answers. The workers were asked, multiple times, to trigger one of these queries. We manually checked the answers for HITs, in order to verify whether workers actually used the chatbot. In addition, conversations were logged and anonymised, and the provided WorkerID was used to track down specific workers and verify the sanity of interaction.

\subsubsection{Sanity check for FS and MFP worker}
To verify that workers actually used the diet apps they were asked to provide a description of the app logo, and to check which particular food (among three alternatives) could be seen on a specified day. As these tasks are subjective and could be failed by legit workers who struggled to use the app, each HIT was manually evaluated to avoid unfair treatment.

\section{Acknowledgements}
This research was funded by the European Union’s Horizon 2020 research and innovation programme under the Marie Skłodowska-Curie grant agreement No 812882.

\newpage
\onecolumn
\section{Appendix A: Quiz sample}
\label{sec:appendix}
\begin{figure*}[!ht]
\centering
\includegraphics[page=1, scale=0.8, trim={0cm 2cm 0cm 2cm},clip]{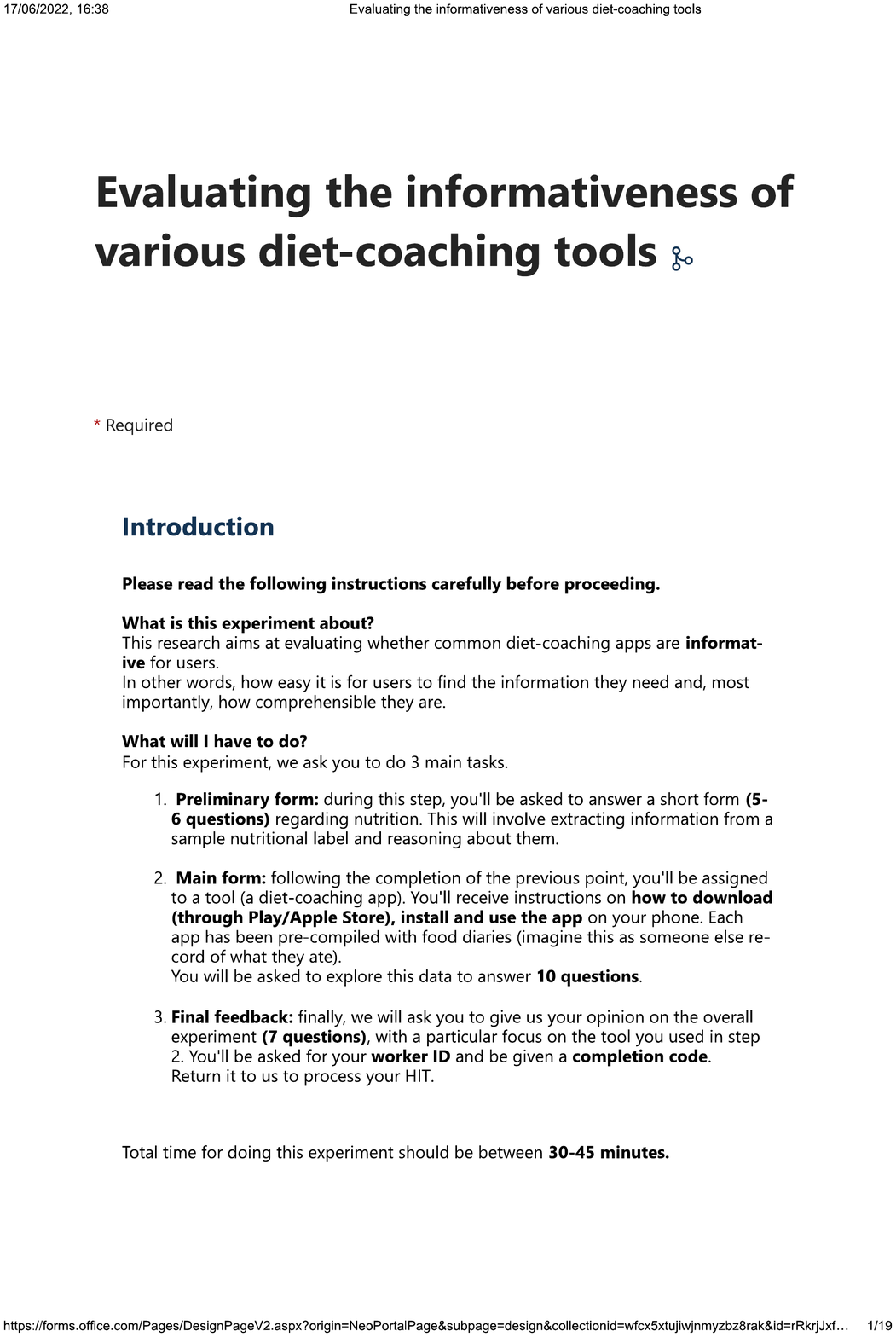}
\end{figure*}
\begin{figure*}
\includegraphics[page=2, scale=0.8, trim={0cm 2cm 0cm 2cm},clip]{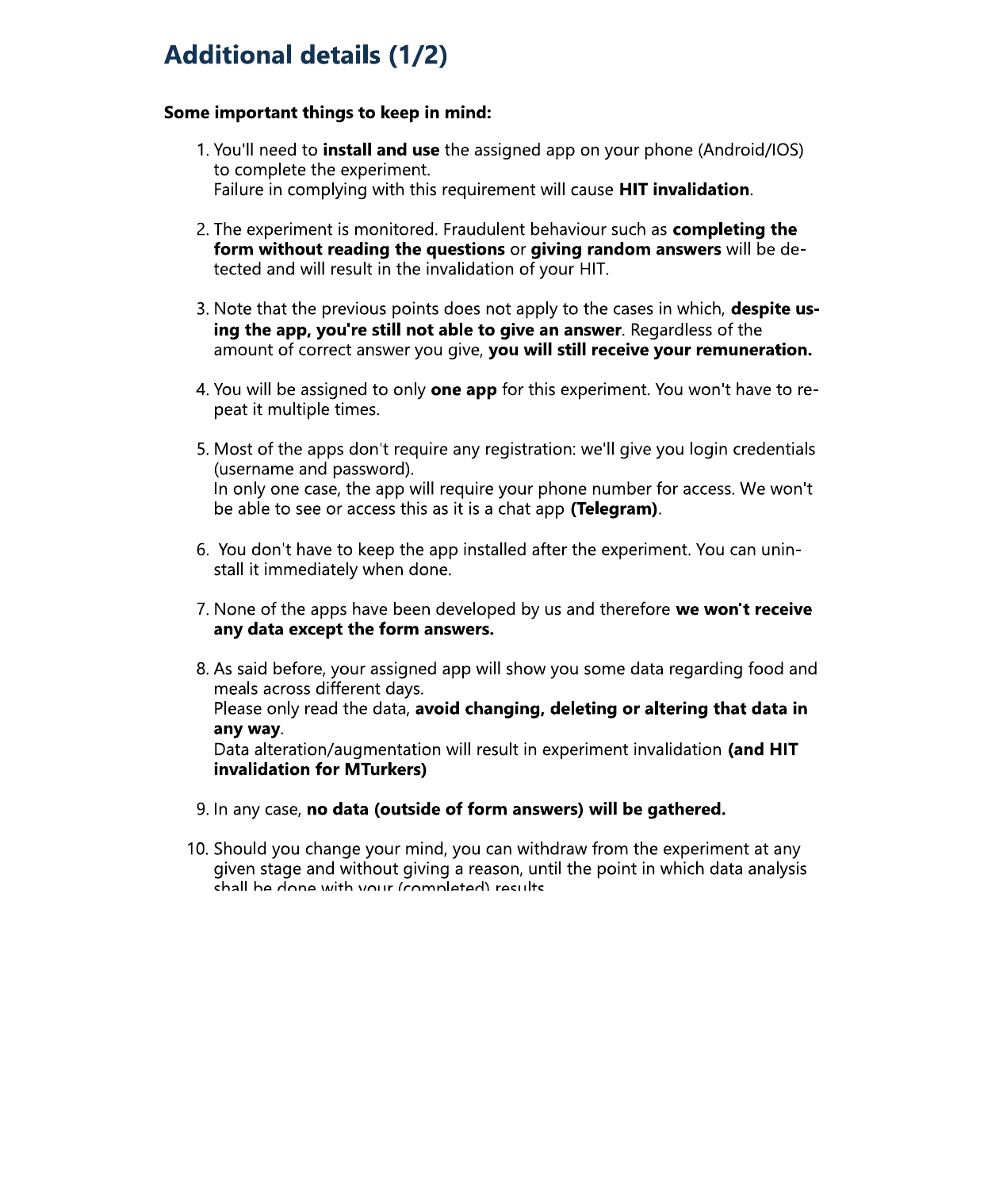}
\end{figure*}
\begin{figure*}
\includegraphics[page=3, scale=0.8, trim={0cm 2cm 0cm 2cm},clip]{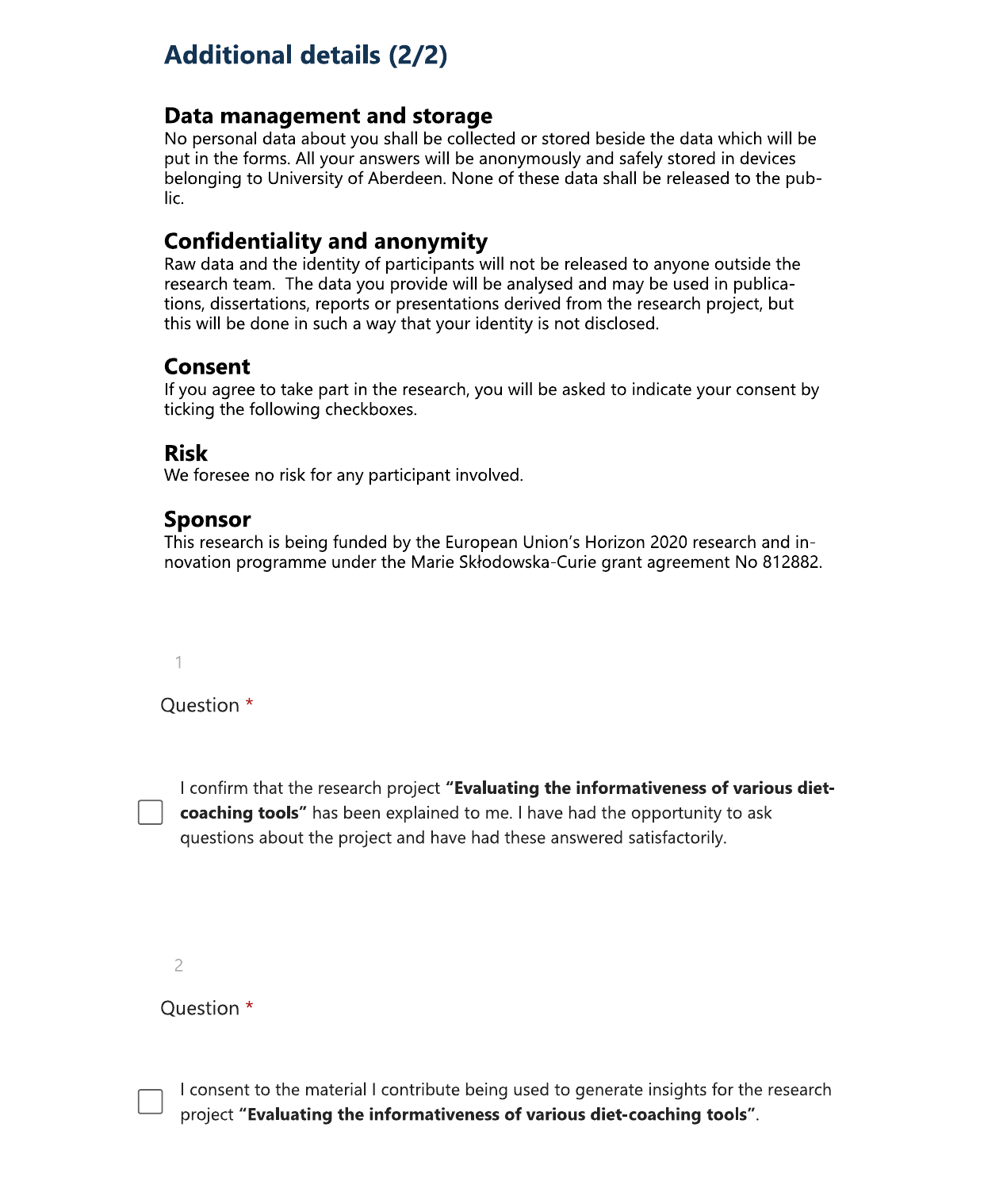}
\end{figure*}
\begin{figure*}
\includegraphics[page=4, scale=0.8, trim={0cm 2cm 0cm 2cm},clip]{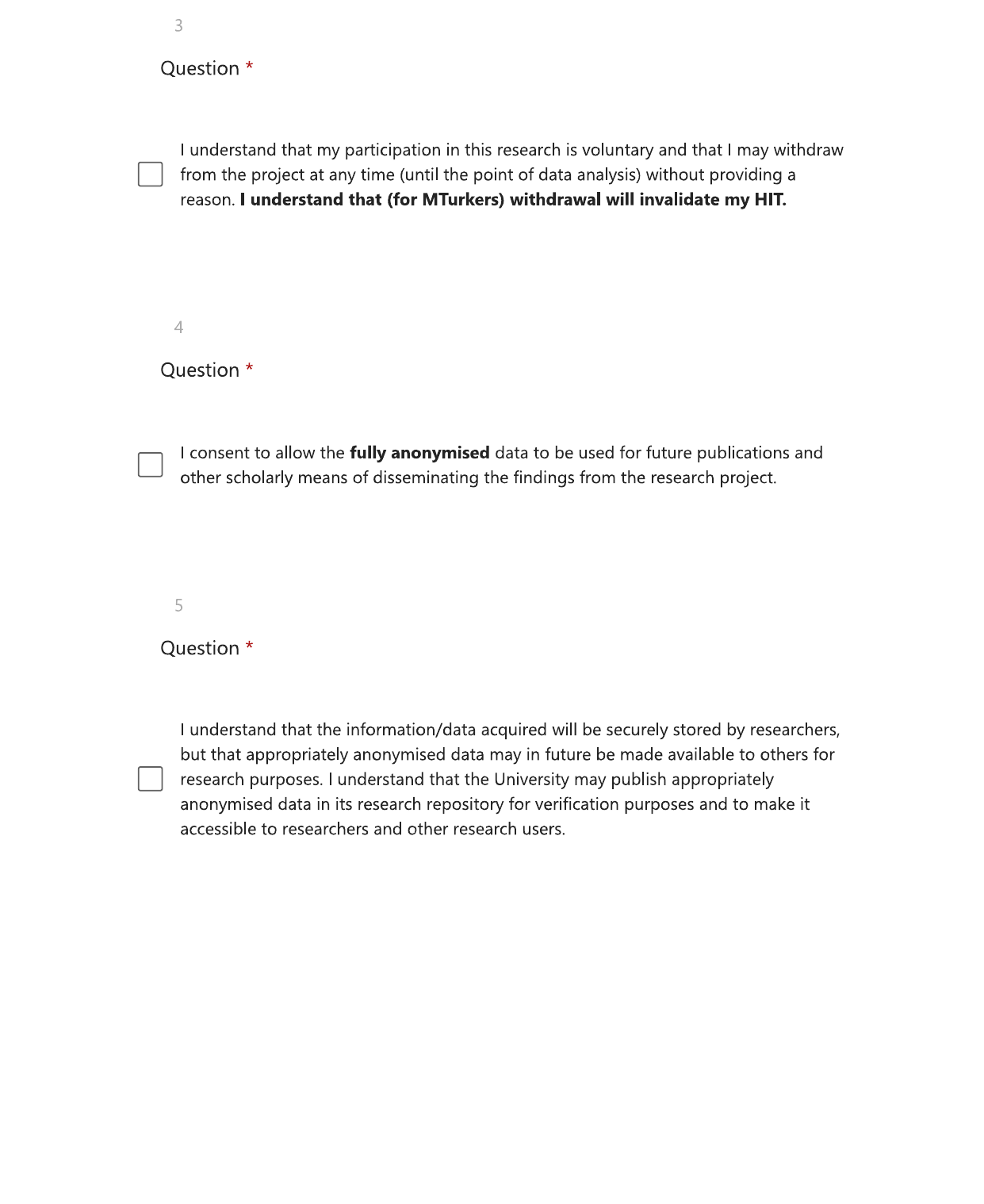}
\end{figure*}
\begin{figure*}
\includegraphics[page=5, scale=0.8, trim={0cm 2cm 0cm 2cm},clip]{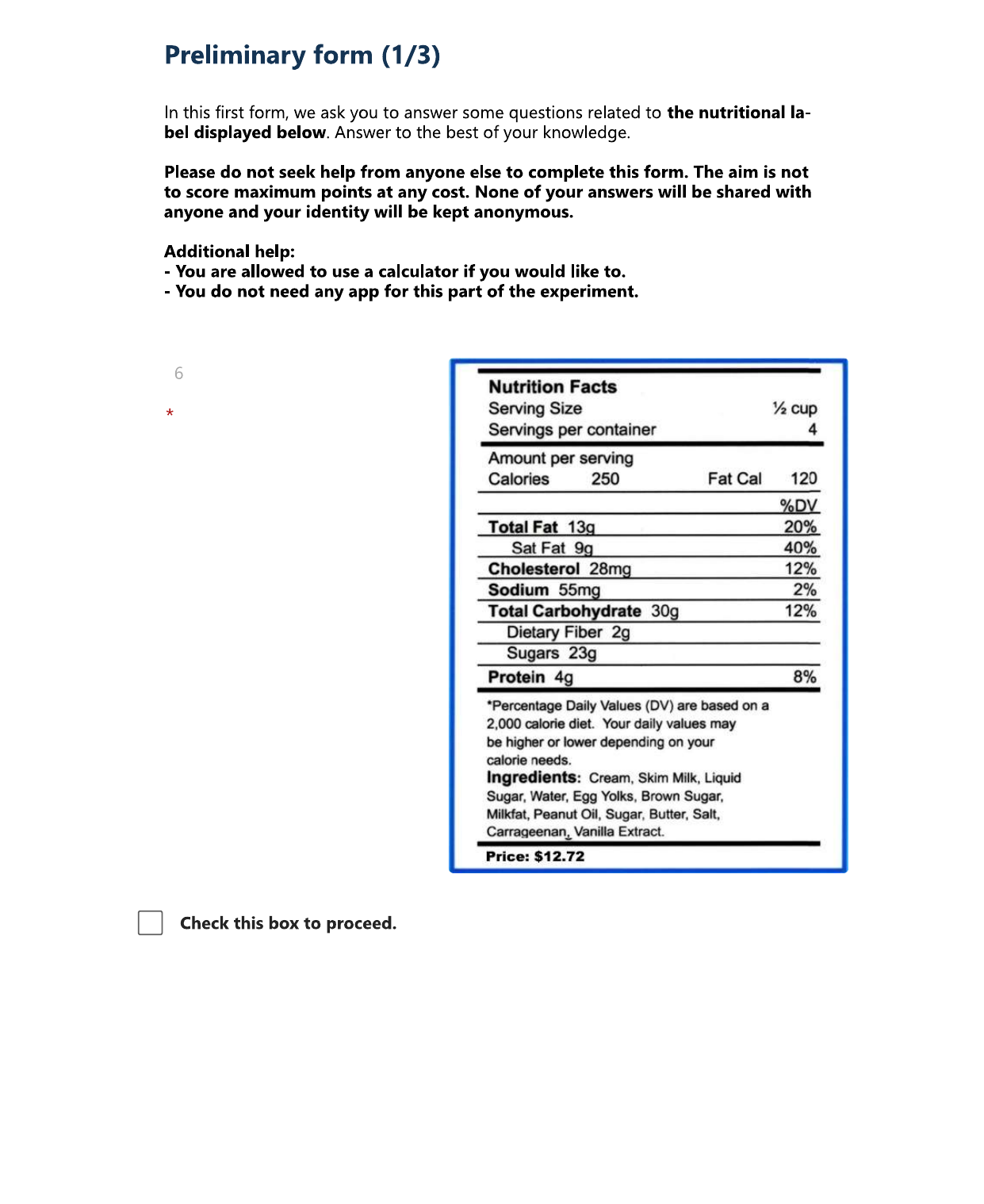}
\end{figure*}
\begin{figure*}
\includegraphics[page=6, scale=0.8, trim={0cm 2cm 0cm 2cm},clip]{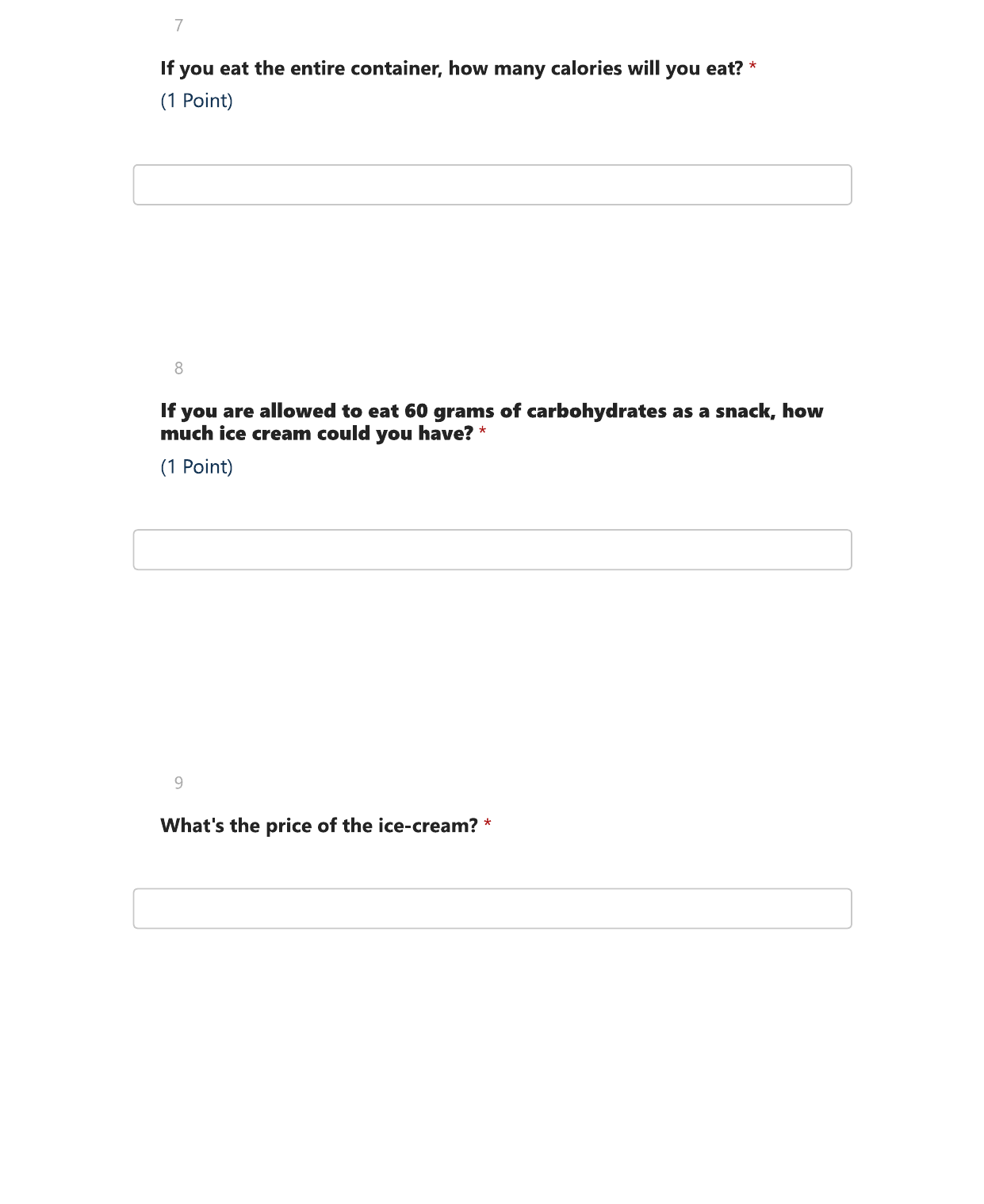}
\end{figure*}
\begin{figure*}
\includegraphics[page=7, scale=0.8, trim={0cm 2cm 0cm 2cm},clip]{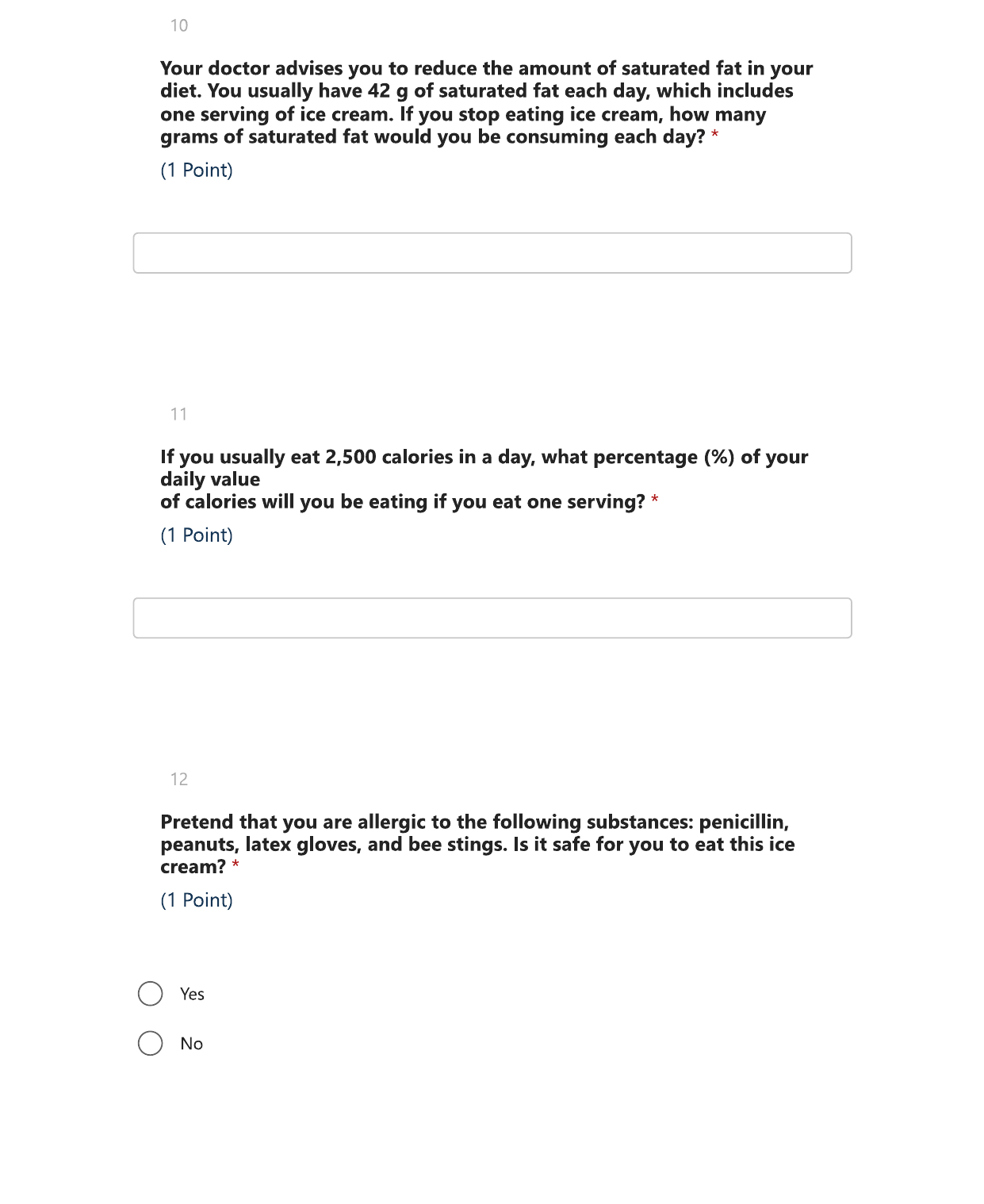}
\end{figure*}
\begin{figure*}
\includegraphics[page=8, scale=0.8, trim={0cm 2cm 0cm 2cm},clip]{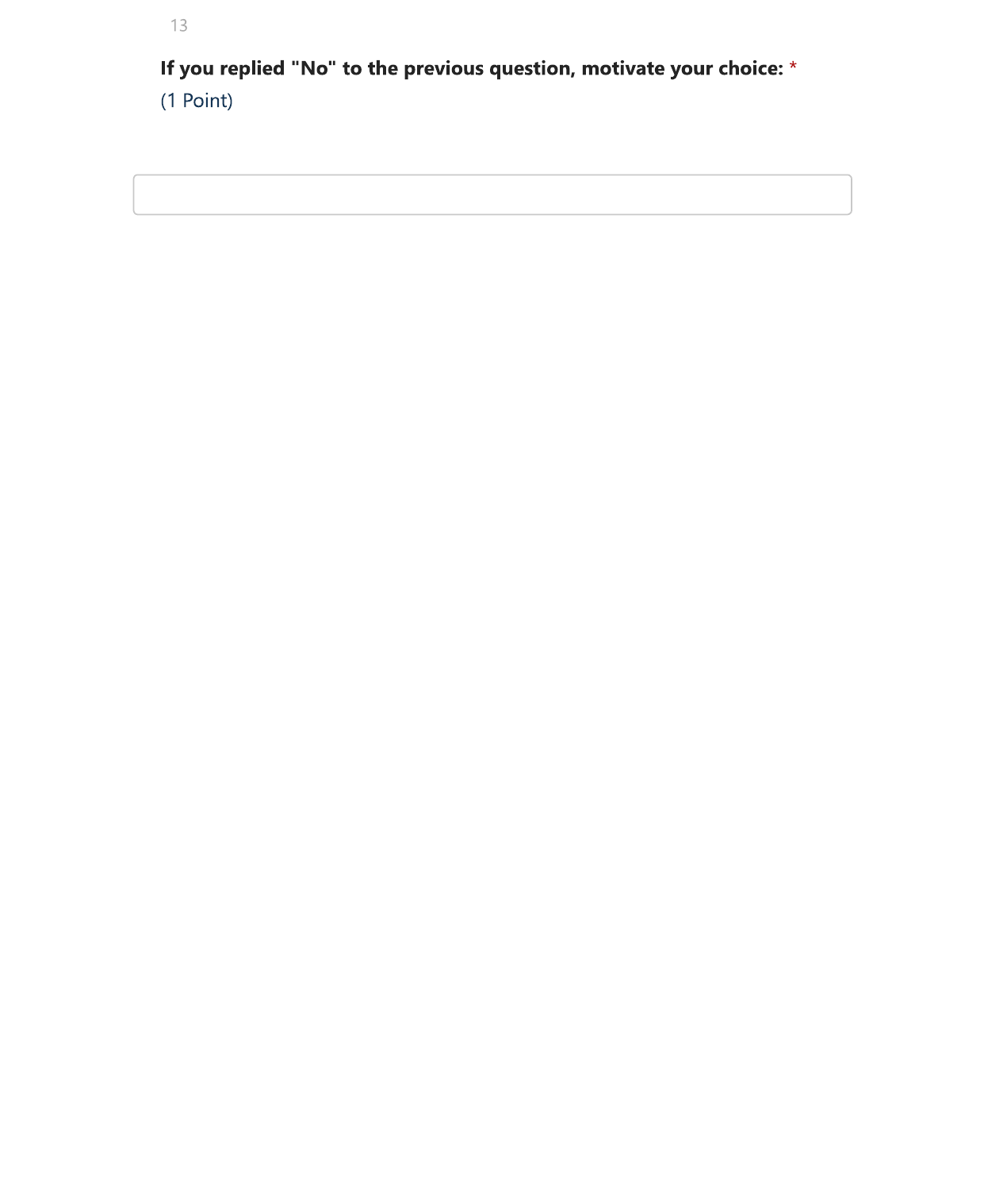}
\end{figure*}
\begin{figure*}
\includegraphics[page=9, scale=0.8, trim={0cm 2cm 0cm 2cm},clip]{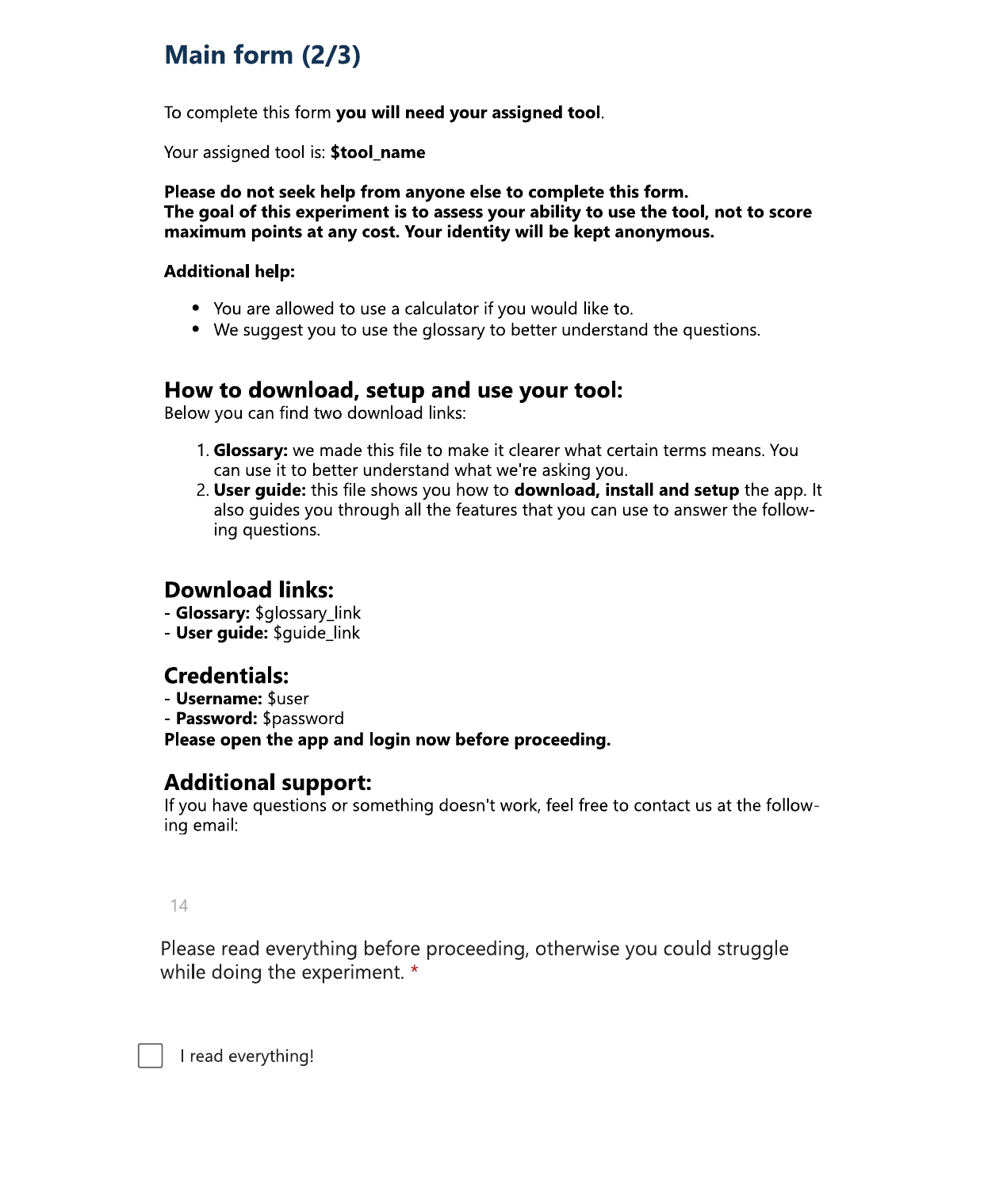}
\end{figure*}
\begin{figure*}
\includegraphics[page=10, scale=0.8, trim={0cm 2cm 0cm 2cm},clip]{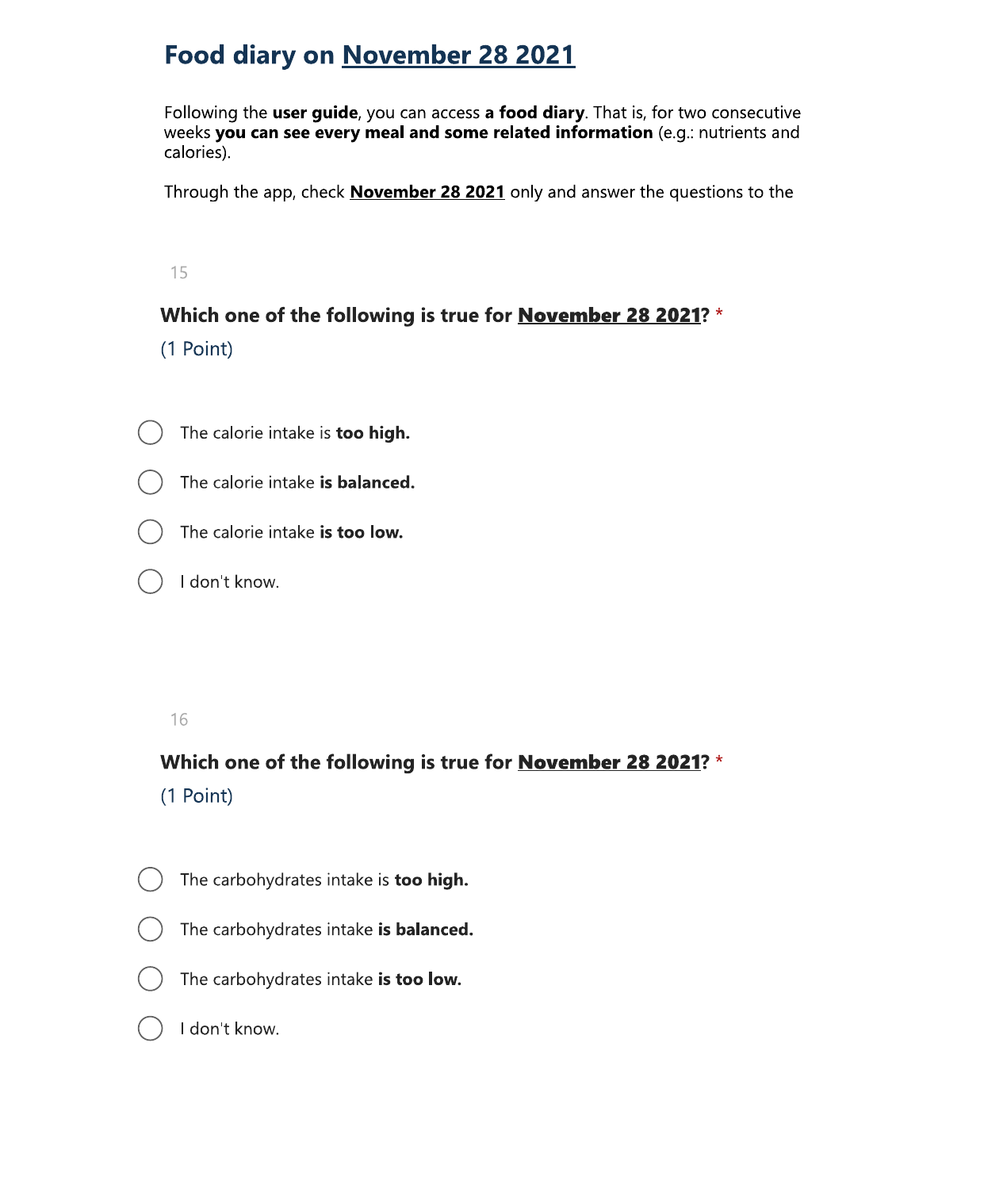}
\end{figure*}
\begin{figure*}
\includegraphics[page=11, scale=0.8, trim={0cm 2cm 0cm 2cm},clip]{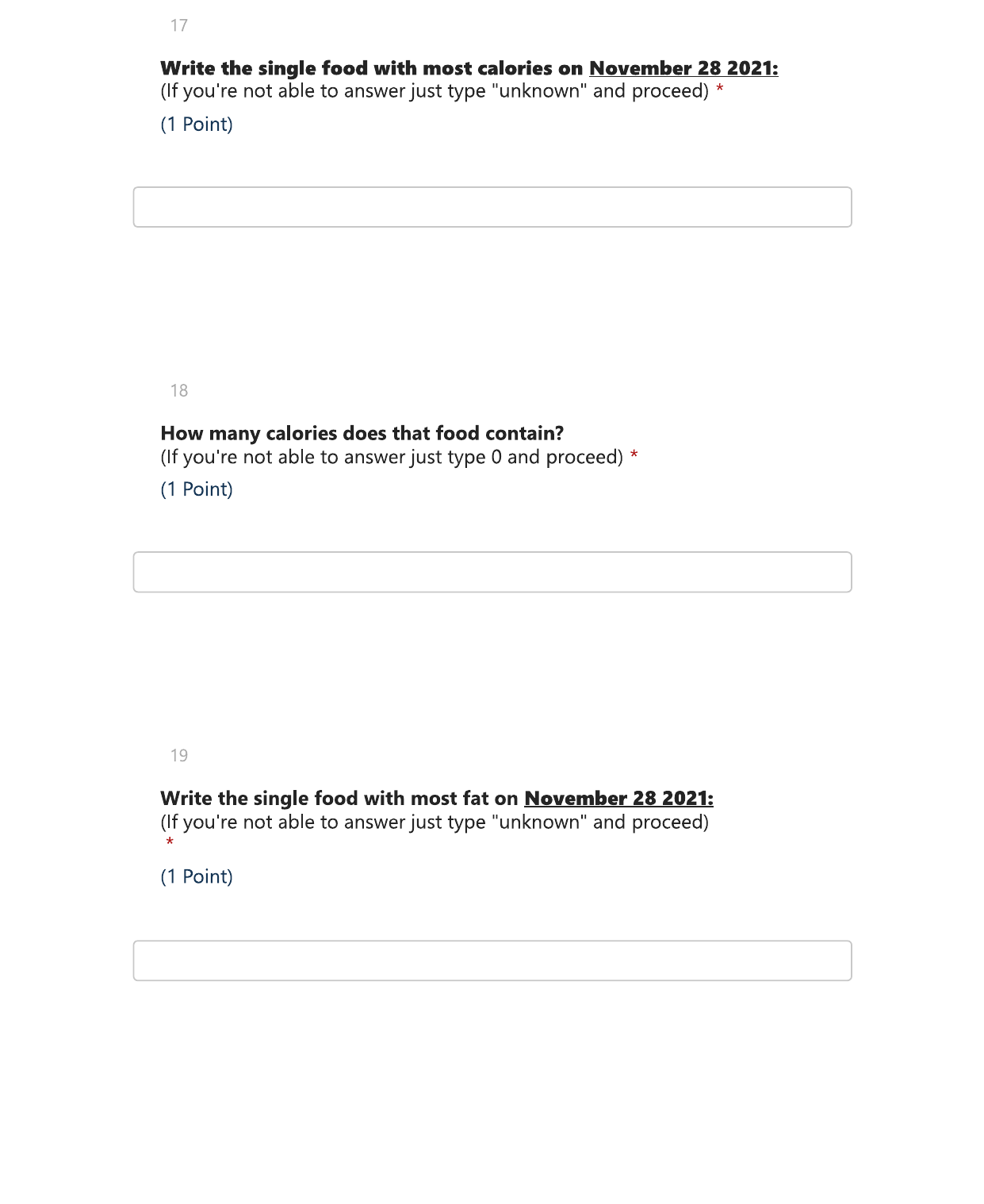}
\end{figure*}
\begin{figure*}
\includegraphics[page=12, scale=0.8, trim={0cm 2cm 0cm 2cm},clip]{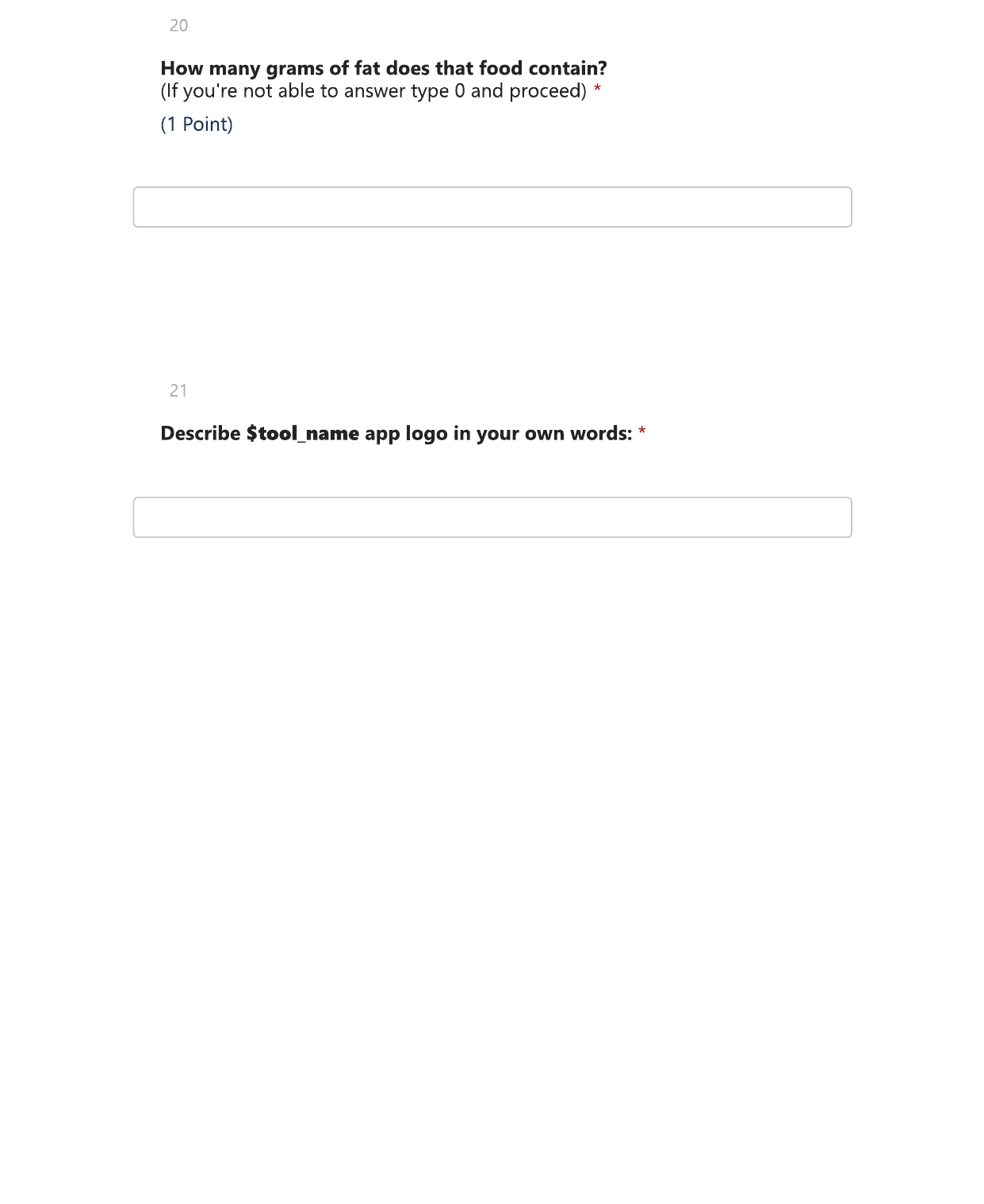}
\end{figure*}
\begin{figure*}
\includegraphics[page=13, scale=0.8, trim={0cm 2cm 0cm 2cm},clip]{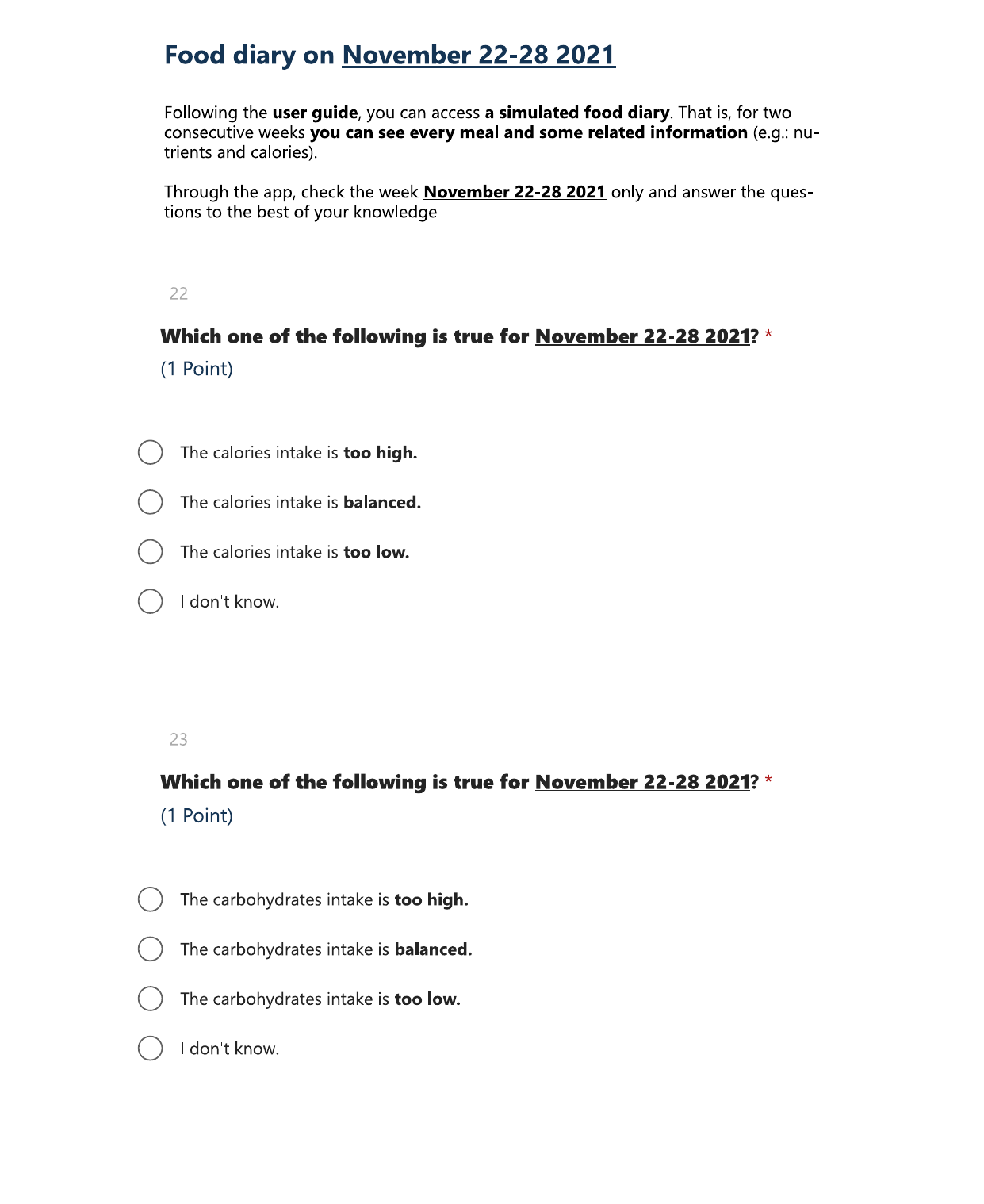}
\end{figure*}
\begin{figure*}
\includegraphics[page=14, scale=0.8, trim={0cm 2cm 0cm 2cm},clip]{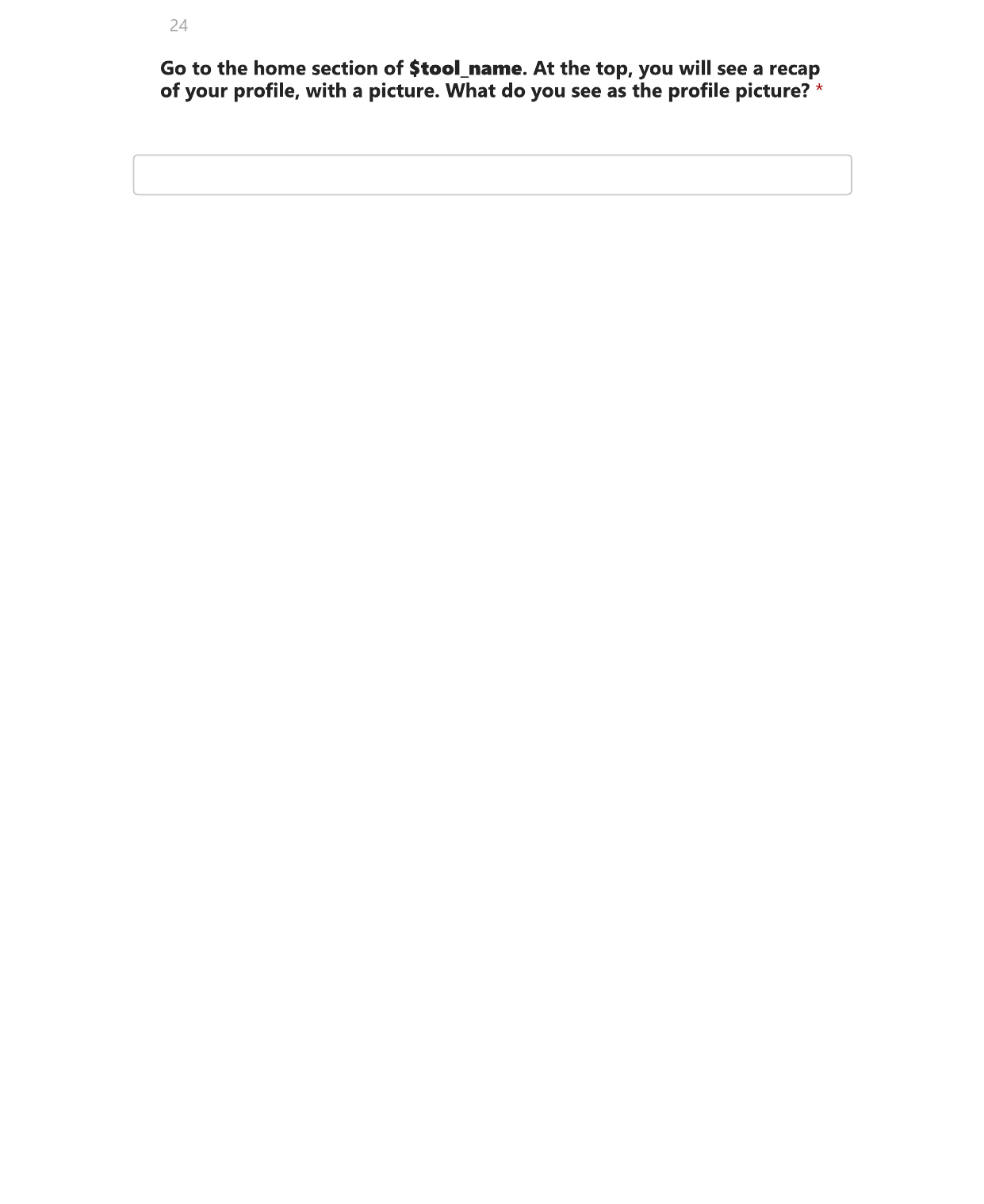}
\end{figure*}
\begin{figure*}
\includegraphics[page=15, scale=0.8, trim={0cm 2cm 0cm 2cm},clip]{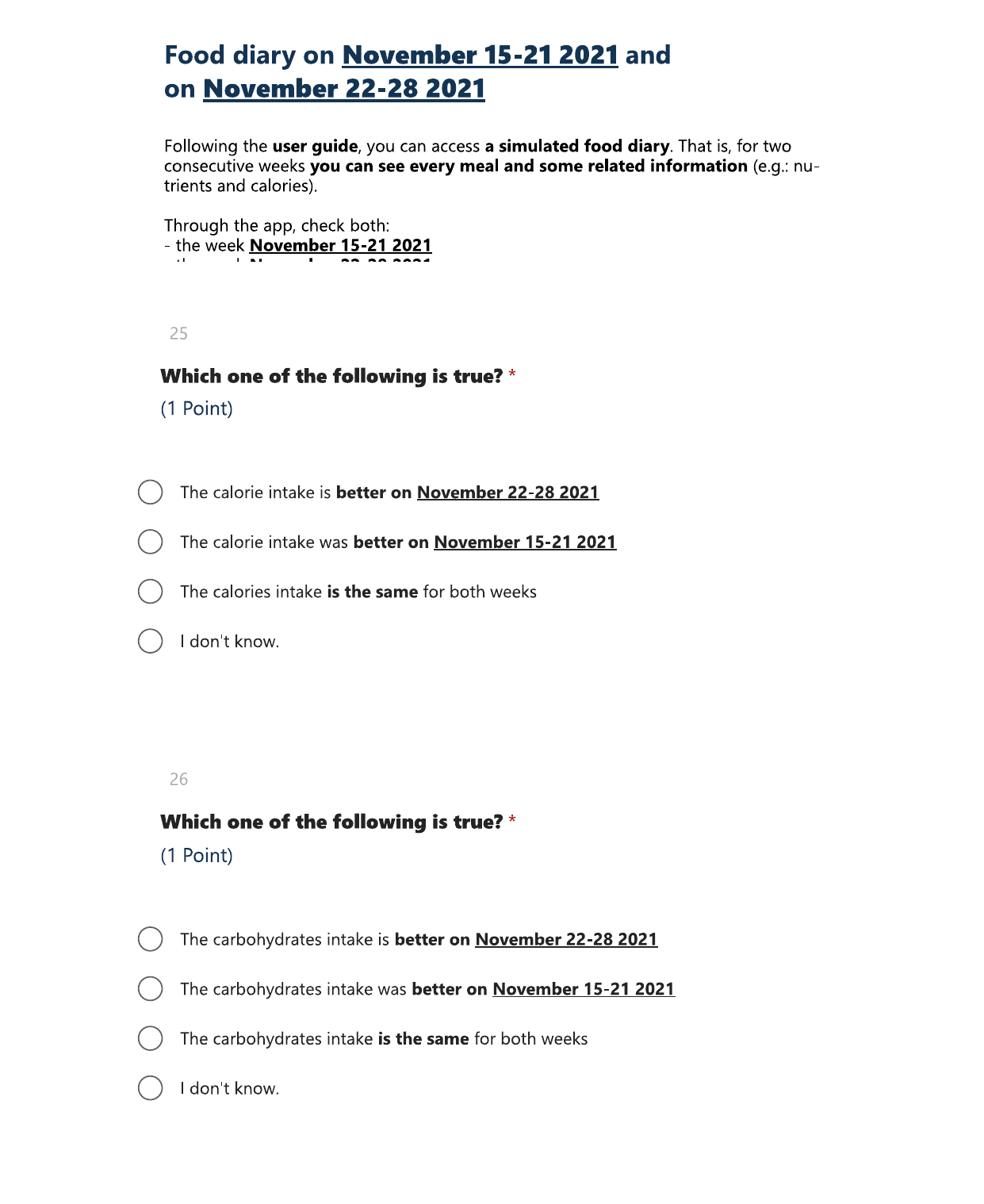}
\end{figure*}
\begin{figure*}
\includegraphics[page=16, scale=0.8, trim={0cm 2cm 0cm 2cm},clip]{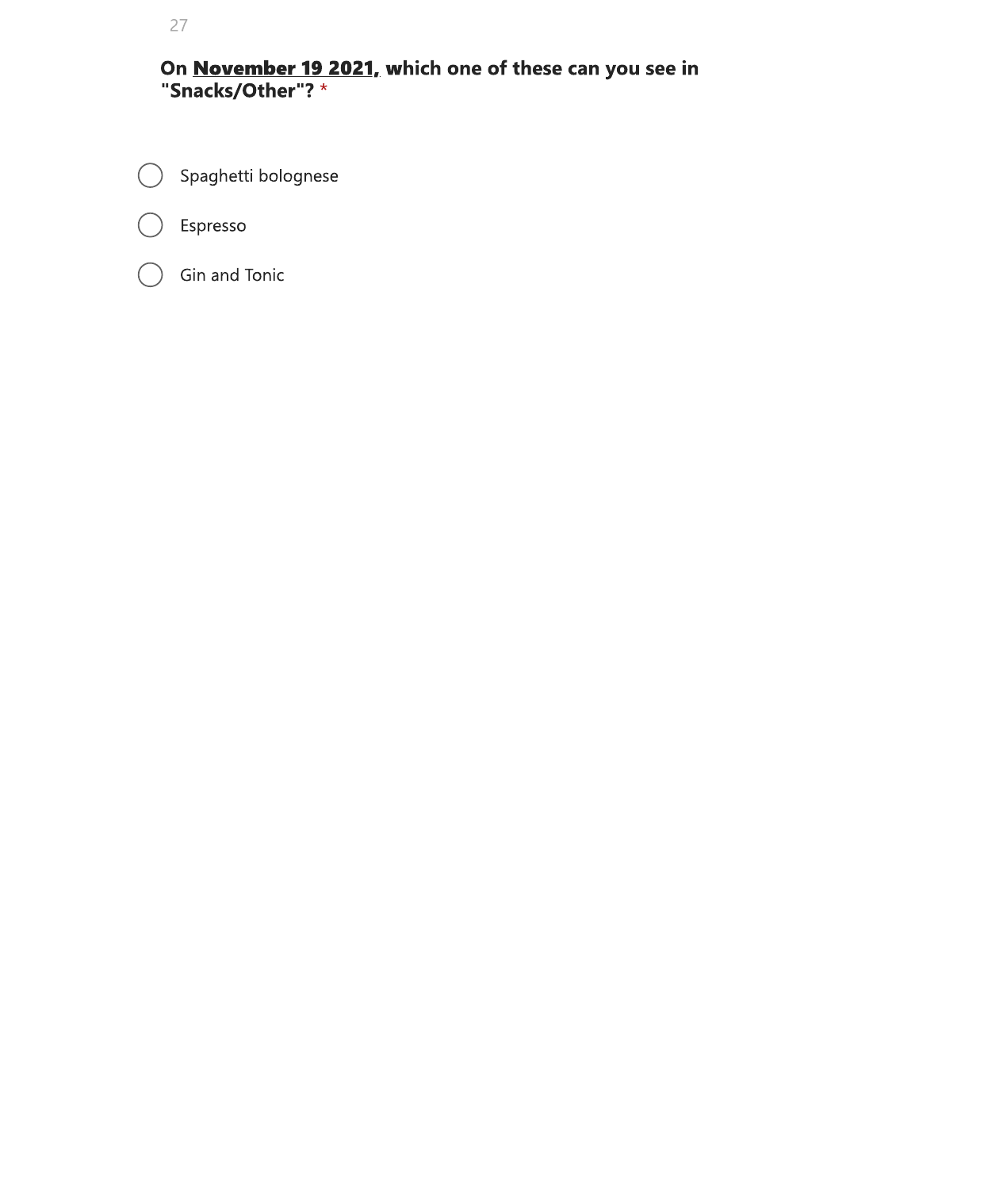}
\end{figure*}
\begin{figure*}
\includegraphics[page=17, scale=0.8, trim={0cm 2cm 0cm 2cm},clip]{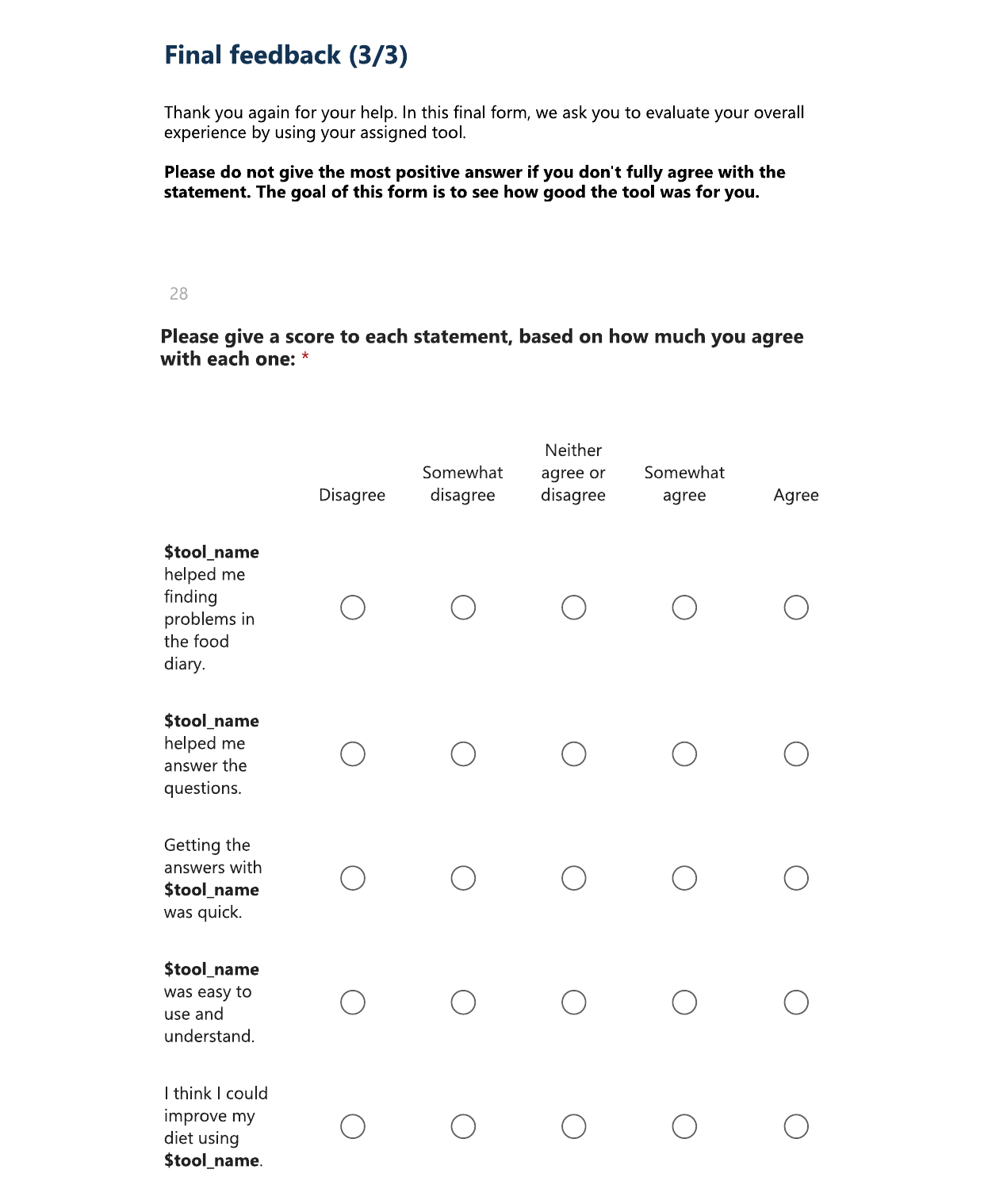}
\end{figure*}
\begin{figure*}
\includegraphics[page=18, scale=0.8, trim={0cm 2cm 0cm 2cm},clip]{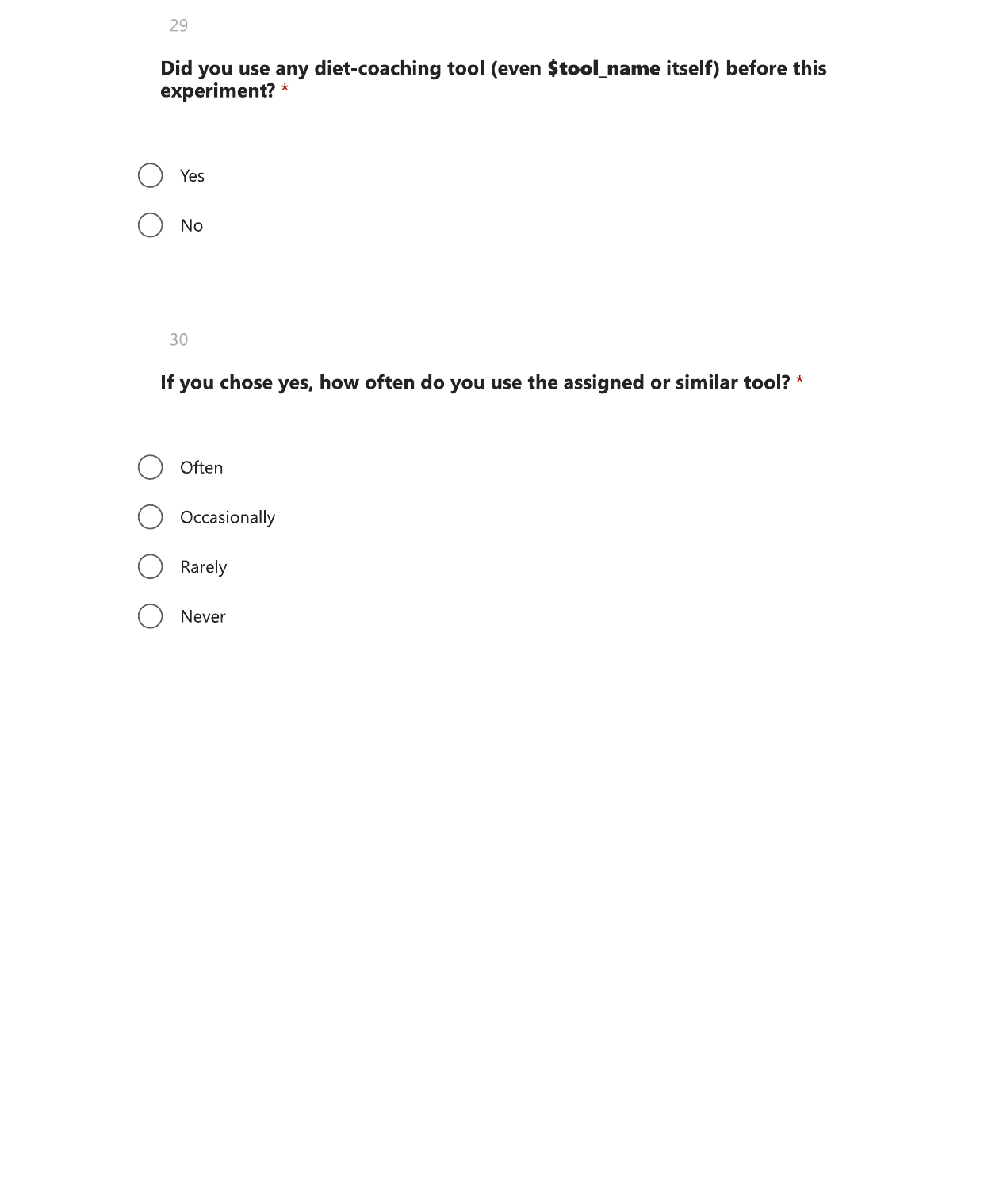}
\end{figure*}

\end{document}